\pdfoutput=1

\documentclass[11pt]{article}

\usepackage[preprint]{acl}

\usepackage{times}
\usepackage{latexsym}
\usepackage{svg}
\usepackage[all]{nowidow}

\usepackage[T1]{fontenc}

\usepackage[utf8]{inputenc}

\usepackage{microtype}

\usepackage{inconsolata}

\usepackage{graphicx}

\usepackage{xcolor}         
\usepackage{colortbl}         
\usepackage{amssymb} 
\usepackage{amsmath}
\usepackage{comment}
\usepackage{booktabs}
\usepackage{multirow}
\usepackage{makecell}
\usepackage{dcolumn}

\usepackage{booktabs}
\usepackage{array}
\usepackage[table]{xcolor}
\usepackage{fontawesome5}
\newcommand{\ghicon}{%
  \raisebox{-0.12em}{\scalebox{1.2}{\faGithub}}%
}

\newcommand{\hficon}{%
  \raisebox{-0.5em}{%
    \includegraphics[height=1.8em]{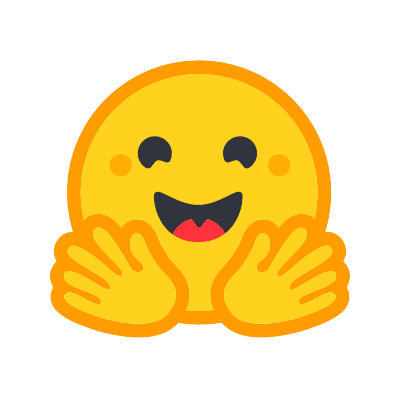}%
  }%
}

\newcommand{\ghlink}[2]{%
  \href{#1}{%
    \ghicon\hspace{0.35em}\texttt{#2}%
  }%
}

\newcommand{\hflink}[2]{%
  \href{#1}{%
    \hficon\hspace{0.35em}\texttt{#2}%
  }%
}

\newcommand{\expected}[1]{\textcolor{green!50!black}{#1}}
\newcommand{\unexpected}[1]{\textcolor{red!70!black}{#1}}
\newcommand{\cd}[2]{$#1_{#2}$}
\newcolumntype{d}[1]{D{.}{.}{#1}}

\usepackage{bm}
\usepackage{tabularx}

\usepackage[nameinlink,capitalize]{cleveref}

\crefformat{section}{\S#2#1#3} 
\crefformat{subsection}{\S#2#1#3}
\crefformat{subsubsection}{\S#2#1#3}
\Crefname{appsec}{appendix}{appendices}
 
\usepackage{enumitem}
\usepackage{threeparttable}
\usepackage{siunitx}

\title{Under the Influence -- of What? \\ A Controlled Evaluation of Influence Functions}

\author{
  \textbf{Johannes Gabriel Sindlinger} \quad
  \textbf{Arzu Burcu Güven} \quad
  \textbf{Rob van der Goot} \quad
  \textbf{Anna Rogers} \\
  IT University of Copenhagen, Denmark \\
  \texttt{\{jgsi, argy, robv, arog\}@itu.dk} \\ [10pt]
  \ghlink{https://github.com/nlpnorth/under-the-influence}{nlpnorth/under-the-influence} \quad
  \hflink{https://huggingface.co/datasets/NLPnorth/under-the-influence}{NLPnorth/under-the-influence}
}

\begin{document}
\maketitle

\begin{abstract}

Influence functions are a promising way to identify which training examples are responsible for a model's output. However, so far empirical evaluations on LLM pretraining level only assumed that they work as intended. This is the first work to systematically evaluate whether the `influential' examples actually have properties relevant to a language model output, based on re-training with controlled data splits. We investigate three prominent Hessian approximation methods (GradSim, TrackStar, K-FAC) in two settings (linguistic phenomena and factual knowledge) in GPT-2 type models (68M-5.6B training tokens). We find that on average, the `influential' examples are indeed relevant, but they do not always bear the target properties directly, and the ratio and kind of relevance varies by setting.

\end{abstract}

\section{Introduction}
\label{sec:introduction}

\begin{figure}[!t]
    \centering
    \includegraphics[
        width=\columnwidth,
    ]{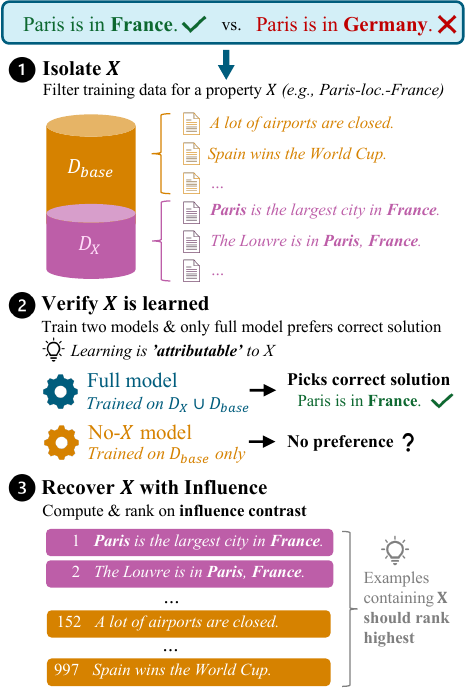}
    \caption{\textbf{Overview of our evaluation framework},
    illustrated on a factual example.}
    \label{fig:overview}
\end{figure}

Understanding which training examples are responsible for a model's behaviour is important for various applications of data attribution, including model debugging, data curation, credit and compensation for the data rights' holders \citep{wuhrlHumanCentricFrameworkData2026}. Among the most widely studied approaches are influence functions (IF), which estimate how a model's predictions would change if a model was retrained without a given training example \citep{kohUnderstandingBlackboxPredictions2017}. Recent work has developed IF approximations scaling to Large Language Models (LLMs) \citep{grosseStudyingLargeLanguage2023, changScalableInfluenceFact2024}.

Despite the principled theoretical grounding, there are also theoretical concerns about what IFs actually reflect \cite{baeIfInfluenceFunctions2022,mlodozeniecDistributionalTrainingData2025}. 
This is compounded by the fact that the empirical evaluations on LLM pretraining data \citep[e.g.][]{changScalableInfluenceFact2024, ruisProceduralKnowledgePretraining2025a} do not explicitly verify that the examples with high influence scores actually have the properties relevant to the model behavior. This matters, because a high score for a query with a property $\mathcal{X}$ may stem from %
a training data point sharing some other feature with the query, which merely co-occurs with $\mathcal{X}$. 

This study is the first to systematically evaluate IFs as an attribution method for LLM pretraining, as shown in \autoref{fig:overview}. First, we \textit{isolate} a property $\mathcal{X}$ by splitting the training corpus into a subset $\mathcal{D}_\mathcal{X}$ containing the property and a base corpus $\mathcal{D}_{\text{base}}$ without it. Then we train a model on $\mathcal{D}$ and a second model only on $\mathcal{D}_{\text{base}}$ to \textit{verify} that $\mathcal{X}$ was learned from $\mathcal{D}_\mathcal{X}$. Finally, we test whether IFs \textit{recover} $\mathcal{D}_\mathcal{X}$ for queries that depend on $\mathcal{X}$. 

We pose three research questions:
\begin{enumerate}[
  leftmargin=2.5em,
  itemsep=0pt,
  topsep=0pt,
  parsep=6pt,
  before=\vspace{0.5\baselineskip},
  after=\vspace{0.5\baselineskip}
]
    \item[\textbf{RQ1}] \textit{Do IFs surface the training data with the target property $\mathcal{X}$?}
    \item[\textbf{RQ2}] \textit{Which types of non‑$\mathcal{X}$ samples are identified as influential?}%
    \item[\textbf{RQ3}] \textit{Which IF approximation most reliably identifies relevant training data?} 
\end{enumerate}

We investigate IFs in two settings: a \textbf{linguistic} setting, where $\mathcal{X}$ is a distinctive grammatical phenomenon, and a \textbf{factual} setting, where $\mathcal{X}$ is a world-knowledge fact. This selection is motivated by prior work, which suggests that influence signals are driven largely by surface-level structure \citep{grosseStudyingLargeLanguage2023} and struggle to recover factual content \citep{changScalableInfluenceFact2024, akyurekTracingFactualKnowledge2022}. 

We find that on average, the examples identified as influential are relevant, but they do not always bear the target properties directly, and the ratio and kind of relevance varies by setting (the kind of target phenomenon, model scale, and IF method). All code to reproduce our experiments and pre-trained models are publicly available.\footnote{\url{https://github.com/nlpnorth/under-the-influence}} We also release a corpus with 10B dependency-parsed tokens of Common Corpus, which we hope would facilitate further work on linguistic phenomena with controlled data splits.\footnote{\url{https://huggingface.co/datasets/NLPnorth/under-the-influence}} 
 
\section{Background}
\label{sec:background}

\subsection{Data Attribution}
Given a model $f_\theta$ trained on a dataset $\mathcal{D}_{\text{train}}$, 
let $\mathcal{L}(z, \theta)$ denote the per-example loss and 
$\phi(\theta) \in \mathbb{R}$ a differentiable observable capturing 
the model behaviour of interest. Data attribution methods aim to assign a scalar score to each training point $z \in \mathcal{D}_{\text{train}}$, reflecting its contribution to $\phi$ 
at the trained parameters $\theta^*$. In most of the previous data attribution work, $\phi$ is chosen as the per-sample loss with respect to a query $z'$, i.e., $\mathcal{L} (\theta^*, z')$, which we follow also in this work.

\paragraph{Influence Functions.}
Among data attribution methods, IFs are particularly prominent, as they offer a principled approach to data attribution grounded in a clear theoretical objective. Originating in robust statistics \citep{hampelInfluenceCurveIts1974} and adapted to deep learning by \citet{kohUnderstandingBlackboxPredictions2017}, they target
the counterfactual \textit{leave-one-out} (LOO) effect: how a model's behaviour changes when a training point $z$ is removed and the model retrained without it. However, realising this in practice proves non-trivial. To compute this directly requires one retraining per training point, which is infeasible at LLM scale.
IFs approximate this counterfactual by
estimating how an infinitesimal downweighting of $z$ shifts the converged
parameters $\theta^*$, and how that shift propagates to
$\mathcal{L}(\theta^*, z')$\footnote{The original formulation of \citet{kohUnderstandingBlackboxPredictions2017} reports the derivative w.r.t. upweighting rather than the leave-one-out estimate itself, yielding a flip of sign in \autoref{eq:if}.}:
\begin{equation}
    \label{eq:if}
    \mathcal{I}(z, z') = \nabla_\theta \, \mathcal{L}(\theta^*, z')^\top
    H_{\theta^*}^{-1} \, \nabla_\theta \mathcal{L}(\theta^*, z)
\end{equation}
where $H_{\theta^*} = \frac{1}{n}\sum_{i=1}^n
\nabla^2_\theta \mathcal{L}(\theta^*, z_i)$ is the Hessian of the average training loss evaluated at the final parameters $\theta^*$.\footnote{Note, that we unit-normalize \autoref{eq:if} following \citet{changScalableInfluenceFact2024} to prevent high-magnitude outlier gradients from dominating the score, shifting \autoref{eq:if} to a cosine-similarity score of preconditioned (and compressed) gradients. See app.~\ref{subsec:appendix_cosine_projections} for details.} See app.~\ref{subsec:appendix_if} for a full derivation of \autoref{eq:if}.

\paragraph{Theoretical Limitations.}
\autoref{eq:if} is well-defined only under assumptions that neural networks usually violate. It assumes a positive-definite Hessian at a true loss minimum. But loss
landscapes are often singular or negatively curved
\citep{sagunEmpiricalAnalysisHessian2018}, and neural models are rarely trained to convergence \citep{baeIfInfluenceFunctions2022}. The standard fixes that restore definiteness have been shown to shift the target away from the LOO effect it is assumed to capture \citep{baeIfInfluenceFunctions2022}. And even computed exactly, the LOO objective itself is increasingly questioned. It is highly sensitive to retraining
stochasticity \citep{mlodozeniecDistributionalTrainingData2025} and prone to missing group-level effects \citep{patelWhenScoreBasedData2026, huMostInfluentialSubset2025}. Hence, there are theoretical reasons why IFs may not work well for deep learning-based LMs. Our work contributes an empirical evaluation of whether they do, based on isolation of target properties they should recover.

\subsection{Approximations}
Independent of any theoretical concerns, computing \autoref{eq:if} is intractable at LLM scale: inverting $H_{\theta^*}$ is prohibitive in both memory and compute \citep{georgeFastApproximateNatural2018, guptaShampooPreconditionedStochastic2018}. Recent work has therefore developed methods to approximate it. %
We test three approaches that differ in how they approximate the curvature $H_{\theta^*}$. We outline the key idea of each below, with full explanations and formulas in app.~\ref{subsec:appendix_curvature}.\footnote{Unrolling-based methods such as MAGIC \citep{ilyasMAGICNearOptimalData2025a} and SOURCE \citep{baeTrainingDataAttribution2024} promise closer alignment to the LOO objective, but their scalability to multi-query settings is limited, so we do not consider them here.\label{fn:unrolling}}

\begin{itemize}[leftmargin=1em, itemsep=2pt, topsep=2pt]
    \item \textbf{Gradient Similarity (GradSim)}, following \citet{pruthiEstimatingTrainingData2020}, applies no curvature correction. It simply sets $\hat{H}_{\theta^*} = I$, and scores a training point by the similarity of its compressed gradient to the query's.
    \item \textbf{TrackStar} \citep{changScalableInfluenceFact2024} approximates the curvature with a block-diagonal pseudo-Hessian, i.e., the autocorrelation matrix of the compressed gradients of both training points and queries. Fitting the autocorrelation matrices requires one pass over the training gradients.
    \item \textbf{(E)K-FAC} \citep{martensOptimizingNeuralNetworks2015, georgeFastApproximateNatural2018} theoretically offers the most faithful approximation of the true curvature. It approximates the Hessian with a Kronecker factorisation of the model's linear layers. \mbox{K-FAC} was originally formulated on full gradients; we adapt it to compressed gradients (see app.~\ref{subsec:appendix_curvature}). As for TrackStar, fitting the `Hessian' requires one pass over training data. Optionally, the factored eigenvalues can be corrected in the Kronecker eigenbasis \citep{georgeFastApproximateNatural2018, grosseStudyingLargeLanguage2023}, yielding EK-FAC at the cost of another training data pass. Since in our experience this correction does not clearly improve attribution (see supplementary results in \autoref{tab:ekfac_ablation}), we use K-FAC only.
\end{itemize}

\section{Related Work}
\label{sec:related}

\paragraph{What do IFs capture at LLM scale?}
\citet{grosseStudyingLargeLanguage2023} first applied IFs  approximation at LLM scale using EK-FAC and observed through qualitative analysis that the signal is driven by surface-level features rather than semantic content. \citet{kowalConceptInfluenceLeveraging2026} show that the standard loss-based influence definition, i.e., selecting the loss on test examples as observable of interest $\phi$ (see \cref{sec:background}), surfaces examples tied to generic concepts in the query rather than the specific target trait, while a concept-based observable recovers it more directly.

\paragraph{IFs for fact retrieval.}
A separate line of work focuses on whether IFs recover the examples supporting factual queries. \citet{changScalableInfluenceFact2024} find that TrackStar underperforms a BM25 baseline at retrieving fact-entailing examples, and \citet{ruisProceduralKnowledgePretraining2025a} report that fact-supporting documents fall within the top-500 in only 30--55\% of cases, using EK-FAC approximation on 7B and 35B models. \citet{akyurekTracingFactualKnowledge2022} fine-tune an mT5 model \citep{xueMT5MassivelyMultilingual2021} on template-based facts it did not already `know', and find that BM25 outperforms both TracIn and embedding-based retrieval.

\paragraph{How do LLMs learn facts?}
Beyond attribution, work on how LLMs acquire facts links factual behaviour to entity (co-)occurrence in the training data. \citet{elazarMeasuringCausalEffects2023} establish this connection causally, and \citet{kandpalLargeLanguageModels2023} show that QA accuracy correlates strongly with entity co-occurrence counts in the pretraining corpus. \citet{nadkarniRewritingHistoryRecipe2026} and \citet{cohenKnowledgelessLanguageModels2026} provide further evidence through interventional retraining, showing that removing or anonymizing entity occurrences suppresses fact learning. However, \citet{nadkarniRewritingHistoryRecipe2026} also find that neither removing co-occurrences nor retrieving relevant documents fully explains fact learning. 
\citet{christophDataKnowledgeEvaluating2025} still observe recall for facts with no matched co-occurrence, which they attribute to answer priors and to transfer from related facts. \citet{zhangCooccurrenceNotFactual2024} show that models are biased to learn co-occurrence statistics rather than factual associations, so co-occurrence in the data does not mean the fact was learned as a fact.

\paragraph{How do LLMs learn linguistic phenomena?}

Most prior work has studied the acquisition of linguistic phenomena through filtered-corpus training, where a phenomenon is removed from the training data and the model is then tested for generalization to it. \citet{misraLanguageModelsLearn2024a} remove article–adjective–numeral–noun (AANN) constructions (e.g. \emph{a beautiful five days}) and find that transformers still generalize to them due to the presence of other related phenomena. \citet{patilFilteredCorpusTraining2024} filter a small (83M-word) Wikipedia corpus for a range of BLiMP phenomena and likewise conclude that models generalize from indirect evidence. Concurrently with our work, \citet{daiWheresIndirectEvidence2026} build on this setup and use IFs to identify what such indirect evidence consists of. For 15 BLiMP phenomena, they find that removing the most influential sentences not matched by the filters lowers accuracy more than removing a random set of equal size. In contrast, \citet{obaCanLanguageModels2024} report a negative result on generalization from indirect evidence. Instead of filtering, they insert nonce words into training as either direct evidence (e.g. \emph{<w> loves herself}) or indirect evidence (e.g. \emph{<w> helping the child loves herself}) for targeted minimal pairs. Models learn from direct evidence but fail to generalize from the indirect cases.

\paragraph{Novelty and contributions.}

To the best of our knowledge, this is the first work to systematically evaluate whether IFs reflect the relevant training data points with specific target properties, as detailed in \cref{sec:methodology}. Prior work on modern generative models does not clearly isolate the target property \citep{changScalableInfluenceFact2024, ruisProceduralKnowledgePretraining2025a, jaburi2025mitigating}.  
Our factual setting shares its design logic with \citet{nadkarniRewritingHistoryRecipe2026} and \citet{cohenKnowledgelessLanguageModels2026}, but they investigated fact acquisition from the perspective of targeted (re-)training, rather than IFs. 
In the linguistic setting, IFs have only been applied in the concurrent work of \citet{daiWheresIndirectEvidence2026}, who use them to discover indirect evidence while the sentences matched by the filters remain in the training data. We instead remove the filtered subset, verify that the phenomenon is learned from it, and test whether IFs recover it, comparing three IF approximations across model scales. To the best of our knowledge, this is also the largest study on neural language models with controlled data ablations for linguistic phenomena to date: we filter up to 5.6B tokens, whereas \citet{misraLanguageModelsLearn2024a,patilFilteredCorpusTraining2024,obaCanLanguageModels2024,daiWheresIndirectEvidence2026} worked on or below 100M scale.


\section{Methodology}
\label{sec:methodology}

We ask whether IFs recover a target property $\mathcal{X}$ embedded in the training data. Our proposed framework requires $\mathcal{X}$ to be isolable by filtering (\emph{distinguishable}) and carried by specific training examples rather than emergent from interactions across them (\emph{independent}).\footnote{This is an idealisation: a model may acquire a property through compositional generalization rather than from individual examples, in which case no filtered $\mathcal{D}_\mathcal{X}$ fully isolates it. We mitigate this by choosing properties that plausibly depend on specific examples (see \cref{sec:experimental-setup}); our verification step then tests whether filtering removes the learned behaviour. See \cref{sec:rq3} for further discussion.}

Let $\mathcal{D} = \mathcal{D}_{\text{base}} \cup \mathcal{D}_\mathcal{X}$ denote the full training corpus, where $\mathcal{D}_\mathcal{X} = \{z_1, \ldots, z_k\}$ is the set of training
examples containing $\mathcal{X}$, and $\mathcal{D}_{\text{base}}$ contains no instance of $\mathcal{X}$. In addition, we require a held-out test set $\mathcal{D}_{\text{test}} = \{(s^+_j, \mathcal{S}^-_j)\}_{j=1}^m$ of \textbf{minimal contrastive examples}, where each $s^+_j$ exhibits $\mathcal{X}$ and every element $s^-$ of $\mathcal{S}^-_j$ is a contrastive counterpart that does not.

\paragraph{Step 1: Isolate $\mathcal{X}$.}
We first localize $\mathcal{X}$ in the training data and filter it out to
obtain the split $\mathcal{D}_{\text{base}} \cup \mathcal{D}_\mathcal{X}$. See \cref{subsec:linguistic_setup} and \cref{subsec:fact_retrieval} for how we operationalize $\mathcal{X}$ in the linguistic and factual settings.

\paragraph{Step 2: Verify that $\mathcal{X}$ is learned.}
We then train a \textbf{full model} $f_\theta$ on
$\mathcal{D}_{\text{train}} = \mathcal{D}_{\text{base}} \cup
\mathcal{D}_\mathcal{X}$, and a \textbf{\mbox{No-$\mathcal{X}$ model}}
$f_{\theta^-}$ on $\mathcal{D}_{\text{base}}$ alone.\footnote{Both models share an initialization (fixed seed per budget). Their training data is ordered independently (app.~\ref{subsec:appendix_models}).} Scoring a sequence $s$ by its summed log-probability $\log p_f(s)$ under a model $f$, we require that
\begin{equation}
\label{eq:model_preference_check}
\begin{aligned}
    \log p_{f_\theta}(s^+_j) &> \log p_{f_\theta}(s^-),
        && \forall\, s^- \in \mathcal{S}^-_j, \\
    \log p_{f_{\theta^-}}(s^+_j) &\leq \log p_{f_{\theta^-}}(s^-),
        && \exists\, s^- \in \mathcal{S}^-_j.
\end{aligned}
\end{equation}
That is, $f_\theta$ should score the positive member above \emph{all} of
its contrastive counterparts, whereas $f_{\theta^-}$ should not: at least
one $s^- \in \mathcal{S}^-_j$ scores at least as high as $s^+_j$. This would
confirm that the learned signal originates from $\mathcal{D}_\mathcal{X}$.
We note that this is a loose criterion for learning $\mathcal{X}$: in
app.~\ref{appendix:metric_verdict} we discuss different scoring metric choices, as well as different definitions of `learning' vs. `not learning', finding no substantive change in our takeaways. 

\paragraph{Step 3: Verify that IFs recover $\mathcal{X}$.}
For a query $s$ and a training point $z \in \mathcal{D}_{\text{train}}$,
let $\mathcal{I}(z, s)$ be the influence of $z$ on $s$. We consider two rankings
of training points:

\emph{Correct-only.} Ranking by $\mathcal{I}(z, s^+_j)$ alone is the standard top-$k$ retrieval task: it returns the training points scored as most influential for the correct sequence.

\emph{Influence contrast.} This setting identifies training samples $z$ with higher influence on the correct than the incorrect test sample \citep{ladhakContrastiveErrorAttribution2023}: 
\begin{equation}
    \Delta \mathcal{I}(z) = \mathcal{I}(z, s^+_j) - \mathcal{I}(z, s^-), \quad s^- \in \mathcal{S}^-_j,
    \label{eq:contrast}
\end{equation}
Since $s^+_j$ and $s^-$ differ only in $\mathcal{X}$, \textbf{the contrast should align most closely with
attribution to $\mathcal{X}$} itself rather than to any other effect. We rank the top-$k$ retrieved training points for each query by the influence contrast. 

In addition to contrastive per-query ranking, we also verify the aggregate condition: whether the contrast is positive for training points
exposing $\mathcal{X}$, and shows no systematic preference for those from
the base corpus:
\begin{equation}
    \label{eq:presence_vs_specifity}
    \begin{aligned}
    \Delta \mathcal{I}(z) &\gg 0 \quad \text{for } z \in \mathcal{D}_\mathcal{X}, \\
    \Delta \mathcal{I}(z) &\lesssim 0 \quad \text{for } z \in \mathcal{D}_{\text{base}}
\end{aligned}
\end{equation}

\section{Experimental Setup}
\label{sec:experimental-setup}

\subsection{Implementation}

\paragraph{Model and training.}
We train models on a range of token budgets, from 68M to 5.6B. For token budgets up to 1.3B we use a GPT-2 small architecture, and GPT-2 medium~\citep{radfordLanguageModelsAre} for larger budgets following the Chinchilla-optimal parameter-to-token ratio
\citep{hoffmannTrainingComputeOptimalLarge2022}. 
Across all budgets we follow Goldfish
\citep{changGoldfishMonolingualLanguage2024} in tokenizer and
hyperparameter setup; \citet{changGoldfishMonolingualLanguage2024} study
language models on small corpora and provide choices calibrated for small
data budgets, addressing shortcomings of the default GPT-2 configuration.
Full
model details and hyperparameters are provided in app.~\ref{subsec:appendix_models}.

As baselines, we use the original GPT-2 small and medium models\footnote{\url{https://huggingface.co/openai-community/gpt2}}, which were trained on approx. 10B tokens of web data sourced from links shared on Reddit \citep{radfordLanguageModelsAre}. 

\paragraph{Attribution methods.}
We evaluate three IF methods introduced in \cref{sec:background}: GradSim, TrackStar, and K-FAC, applied with the \textsc{Bergson} library \citep{quirkeBergsonOpenSource2026}. The baseline is BM25 \citep{robertsonOkapiTREC31994}, as implemented by~\citet{luBM25SOrdersMagnitude2024}.\footnote{BM25 is the standard baseline for the factual setting \cite{changScalableInfluenceFact2024, akyurekTracingFactualKnowledge2022}, and we retain it for the linguistic setting, where there is no comparable work. However, due to the nature of the linguistic contrastive pairs, for BM25 we retain stopwords, which are normally discarded. Some minimal pairs differ only in a function word (e.g. negation), and if they were stripped, $s^+$ and $s^-$ would have identical
token sets.}

Scoring the full corpus is computationally expensive, so we subsample
before attribution, following the approach of
\citet{ruisProceduralKnowledgePretraining2025a}.\footnote{We use the same subsample for both attribution steps, i.e., fitting the Hessian approximation and scoring.} In the linguistic setting
we draw $N = 5{,}000{,}000$ sentences while preserving the
full-corpus ratio $|\mathcal{D}_\mathcal{X}|/|\mathcal{D}_{\text{base}}|$. In the factual setting attribution operates at the level of an
individual fact, so preserving the ratio would require a separately sampled
index per fact. We therefore fix one base sample of $5{,}000{,}000$
sentences and add $\mathcal{D}_\mathcal{X}$ for each fact on top.
 
\subsection{Linguistic Setting}
\label{subsec:linguistic_setup}

In this setting, $\mathcal{X}$ is a grammatical phenomenon, and $\mathcal{D}_\mathcal{X}$ consists of training sentences exhibiting it. We study four phenomena:
\begin{itemize}[leftmargin=1em, itemsep=0pt, topsep=4pt]
\item \emph{Existential there quantifiers} \footnotesize \\ 
    E.g.: \expected{\emph{There are few cars.}} vs. \unexpected{\emph{*There are all cars.}}
\item \normalsize \emph{Binding reflexives} \footnotesize \\ 
    E.g.: \expected{\emph{The boy hurt himself.}} vs. \unexpected{\emph{*The boy hurt herself.}} 
\item \normalsize \emph{Wh-island effects} \footnotesize \\ 
    E.g.: \expected{\emph{Which girls ran?}} vs. \unexpected{\emph{*Which ran girls?}}
\item \normalsize \textit{Licensed negative polarity items (NPIs)} \footnotesize \\
    E.g.: \expected{\emph{I'm not playing anymore.}} vs. \unexpected{*\emph{I'm playing anymore.}} 
\end{itemize}

We choose these four phenomena as we hypothesize that they meet our \emph{independent} and \emph{distinguishable} requirements on
$\mathcal{X}$ (\cref{sec:methodology}): they represent unique functions within the language, that generally depend on closed sets or rare sentence types. Binding reflexives and NPIs each depend on specific closed-class forms (e.g., \emph{himself}, \emph{any}). Wh-island effects are restricted to interrogative constructions, which are rather infrequent in text corpora. Existential there constructions are frequent in general, but restricting to quantifier probes yields a closed, well-defined set of function words. 

The main evaluation dataset is BLiMP \cite{warstadtBLiMPBenchmarkLinguistic2023}, except for NPIs, where we observed issues with the BLiMP data. For this phenomenon, we use a subset of BLiMPPI dataset \cite{güven2026blimppibenchmarklinguisticminimal} (see~app.~\ref{subsec:appendix_blimp_npi} for details).

Our models are trained on a sample (up to 5.6B tokens) from the English part of Common Corpus \citep{langlaisCommonCorpusLargest2026}: a collection of public domain and permissively licensed texts. For the linguistic filtering, we annotate the data by training a MaChAmp parser~\cite{van-der-goot-etal-2021-massive} on the multi-domain GUM corpus~\cite{zeldes2017gum} with deberta-v3-large~\cite{he2023debertav} as backbone (more details about the corpus and parsing are available in app.~\ref{subsec:appendix_corpora}). For each phenomenon, we apply a corresponding syntactic filter over the parsed corpus, detailed in app.~\ref{appendix:filters}.

\subsection{Factual Setting}
\label{subsec:fact_retrieval}
We define $\mathcal{X}$ as a world-knowledge fact expressed as a relation between two entities (e.g., \emph{Warsaw--capital--Poland}). We use the BEAR probing benchmark \citep{wilandBEARUnifiedFramework2024}, which consists of $7{,}731$ such relations implemented as template-based multiple-choice problems, with slots filled by candidate entities. Each relation is expressed via three templates (e.g., \emph{X was born in Y}; \emph{X's place of birth is Y}; \emph{The birthplace of X is Y.}). Only one of the entities (Y) is considered as a target, with 5-60 pre-defined candidates (avg.\ 31.7), ranked by completion likelihood.

Preliminary experiments showed that our subset of Common Corpus provides limited support for factual information in our target benchmark.\footnote{Only 601 of the 7{,}731 entity-relation triplets in BEAR \citep{wilandBEARUnifiedFramework2024} have more than 10 co-occurrences count of in the 5.6B corpus. The resulting model achieves only 3.6\% accuracy on BEAR, compared to 7.1\% and 10.1\% for GPT-2-small and GPT-2-medium (see \autoref{tab:facts_accuracy}).}
We therefore train an additional model on an English Wikipedia dump \citep{wikimedia_english_wikipedia_dump_2023_11_01}, following \citet{christophDataKnowledgeEvaluating2025}. We use the Goldfish tokenizer, trained on the Common Corpus subset, for both corpora. The Wikipedia corpus was tokenized to 4.8B tokens.

Even when training on Wikipedia, our goal is not to test the model for overall factual knowledge, but only to see whether it learned those facts that it had a fair chance to learn. Hence we restrict the analysis to \textit{corpus-supported facts} with at least 10 entity co-occurrences in the training corpus: below this, we expect the model to have too little evidence to plausibly learn the fact. For each corpus-supported fact, which the full model $f_\theta$ has learned, we filter the training corpora by collecting every sentence where the two entities of a fact co-occur (aliases and demonyms resolved). Following \citet{christophDataKnowledgeEvaluating2025}, we additionally filter out sentences that mention only one of the two entities, provided that this entity has fewer than 30,000 mentions in the corpus. We do not apply this to more frequent entities (e.g., \emph{France}), since removing all their mentions would delete a large share of the corpus and shift its overall distribution.

\begin{table}[t]
  \centering
  \footnotesize
  \setlength{\tabcolsep}{2.1pt}
  \begin{tabular}{l r@{}l r@{}l r@{}l r@{}l}
    \toprule
    & \multicolumn{2}{c}{\makecell{Bind.\\Refl.\\{\tiny $N{=}1{,}000$}}}
    & \multicolumn{2}{c}{\makecell{NPI\\\vphantom{Refl.}\\{\tiny $N{=}100$}}}
    & \multicolumn{2}{c}{\makecell{Exist.\\\emph{there}\\{\tiny $N{=}1{,}000$}}}
    & \multicolumn{2}{c}{\makecell{Wh-\\Island\\{\tiny $N{=}1{,}000$}}} \\
    \midrule
    68M   & 58.1 & \cd{}{\unexpected{+3.6}}          & 34.0 & \cd{}{\expected{-16.0}}
          & 91.7 & \cd{}{\expected{\bm{-23.2}}}      & 21.8 & \cd{}{\expected{-9.2}} \\
    130M  & 59.6 & \cd{}{\expected{-2.7}}            & 41.0 & \cd{}{\expected{-15.0}}
          & 97.1 & \cd{}{\expected{-9.6}}            & 22.0 & \cd{}{\expected{-2.5}} \\
    \cmidrule(lr){1-9}
    1.3B    & 74.7 & \cd{}{\expected{\bm{-44.2}}}      & 73.0 & \cd{}{\expected{-27.0}}
          & 98.1 & \cd{}{\expected{-4.7}}            & 39.3 & \cd{}{\expected{-11.5}} \\
    5.6B  & 75.0 & \cd{}{\expected{-25.1}}           & 86.0 & \cd{}{\expected{\bm{-32.0}}}
          & 98.1 & \cd{}{\expected{-6.8}}            & 58.7 & \cd{}{\expected{\bm{-14.2}}} \\
    \midrule
    {\scriptsize \textit{GPT-2-small}}  & 67.3 && 87.8 && 97.0 && 57.7 & \\
    {\scriptsize \textit{GPT-2-medium}} & 68.6 && 87.0 && 98.5 && 50.4 & \\
    \midrule
    {\scriptsize \textit{Random Guessing}} & 50.0 && 50.0 && 50.0 && 50.0 & \\
    \bottomrule
  \end{tabular}
\caption{\textbf{Model performance on linguistic phenomena across token
  budgets.} Accuracy (\%) on minimal pairs; the main number is the model trained
  on the full corpus, the subscript is
  the change from filtering $\mathcal{X}$. \expected{Green (-)} marks the drop, \unexpected{red (+)} the one case here where filtering \textit{increased} accuracy. $N$ is the number of minimal pairs. Full results are available in app.~\autoref{tab:filtering_linguistic_full}. }
  \label{tab:filtering_linguistic}
\end{table}

\section{Results}
\label{sec:results}

\subsection{Verifying that the models learned $\mathcal{X}$}
\label{sec:results_learned}

\paragraph{Linguistic phenomena.}
Despite smaller token budgets, our models are competitive with the GPT-2 baselines across all four phenomena (\autoref{tab:filtering_linguistic}). Accuracy rises with budget, and so does the drop in accuracy under filtering: at 1.3B and 5.6B, ablating $\mathcal{X}$ pushes accuracy down to (or below) chance for every phenomenon except \textit{existential there quantifiers}. We hypothesize that this is due to filters not catching some noisy data points. This phenomenon also proves to be overall `easy' to learn, which would explain that the impact of filtering is high only for the small token budgets.

\paragraph{Facts.}
Both baselines and our models score poorly on the full $7{,}731$ BEAR relations (\autoref{tab:facts_accuracy}).\footnote{The default BEAR setup is to average across all templates, disregarding the fact level. Supplementary evaluation in this setting is provided in \autoref{fig:standard-bear-results}.} For our models, the likely explanation is that most relations are not corpus-supported (which we define as $\geq 10$ entity co-occurrences): only $601$ relations qualify in the Common Corpus subset, and $3{,}221$ in the Wikipedia corpus. Our model trained on Wikipedia reaches $13.4\%$, exceeding the GPT-2 baselines despite far fewer training tokens. As discussed in \cref{subsec:fact_retrieval}, since our goal is to establish whether the models learn the facts that were available to them, we focus only on the part of BEAR that is supported by the training data.

\begin{table}[t]
  \centering
  \footnotesize
  \setlength{\tabcolsep}{8pt}
  \begin{tabular}{l r r@{}l r@{}l}
    \toprule
    & \multicolumn{1}{c}{%
        \makecell{All\\facts\\{\tiny $N{=}7{,}731$}}}
    & \multicolumn{2}{c}{%
        \makecell{CC-\\supported\\{\tiny $N{=}601$}}}
    & \multicolumn{2}{c}{%
        \makecell{Wiki-\\supported\\{\tiny $N{=}3{,}221$}}} \\
    \midrule
    CC 5.6B
      & 3.6
      & 18.8 & \cd{}{\expected{-3.7}}
      & \multicolumn{2}{c}{--} \\

    Wikipedia 4.8B
      & 13.4
      & \multicolumn{2}{c}{--}
      & 20.5 & \cd{}{\expected{-3.2}} \\
    \midrule
    {\scriptsize \textit{GPT-2-small}}
      & 7.1 & 28.1 && 11.6 & \\

    {\scriptsize \textit{GPT-2-medium}}
      & 10.1 & 39.6 && 18.3 & \\
    \midrule
    {\scriptsize \textit{Random Guessing}}
      & 0.0\rlap{4} & 0.1 && 0.0\rlap{4} & \\
    \bottomrule
  \end{tabular}
  \caption{\textbf{Model performance on factual knowledge (BEAR).}
  Accuracy (\%) per fact, where a fact counts as learned only if the correct
  object ranks highest under \emph{all three} of its templates. 
  \emph{All facts:} the full benchmark. \emph{CC-} /
  \emph{Wiki-supported:} facts whose entity co-occurrence is $\geq 10$ in
  the respective corpus. Subscripts give
  $\Delta=\text{No-}\mathcal{X}-\text{Full}$
  (\expected{green (-)}: the drop), reported only where a filtered
  counterpart exists, i.e., each model on its own corpus's subset.
  Full results available in app.~\autoref{tab:filtering_facts_full}.}
  \label{tab:facts_accuracy}
\end{table}

\begin{figure*}[!t]
    \centering
    \includegraphics[width=\linewidth]{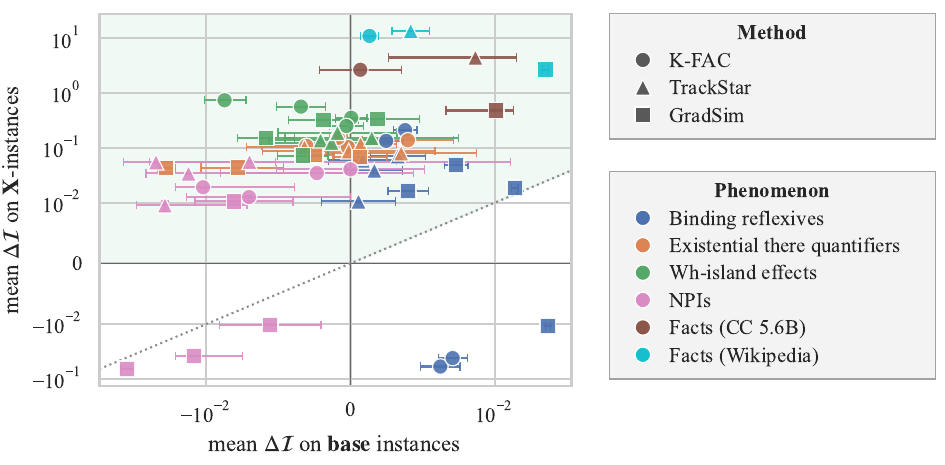}
\caption{
\textbf{Do IFs surface the training data with property $\mathcal{X}$? In aggregate, yes.} Each point compares a method's mean influence contrast on $\mathcal{X}$-instances (y-axis) against base instances (x-axis), for one phenomenon $\times$ method $\times$ token budget. The large majority sit in the \textcolor[HTML]{16a34a}{shaded} region above the $y{=}x$ line, satisfying \autoref{eq:presence_vs_specifity} (see footnote~\ref{fn:implementation_aggregated_contrast} for details). Bars are 95\% CIs (negligible on the y-axis).
}
\label{fig:presence_vs_specifity}
\end{figure*}

In that setting, we find that filtering the fact-supporting data has a much smaller effect compared to the linguistic setting: the full-corpus-supported accuracy is reduced only by $3.7$ points (CC 5.6B) and $3.2$ (Wikipedia). Only $52$ (CC 5.6B) and $261$ (Wikipedia) facts are not learned by the filtered-data model, and learned by the full model. Filtering even \emph{helps} the filtered-data model with some relations where the full model is wrong, which we attribute to cross-relational interference in the multi-choice evaluation. For example, removing `Warsaw' eliminates it as a distractor in other country-capital queries, shifting mass to the correct object (e.g., letting `Prague' win for `Czech Republic' where the full model failed).

\emph{In the subsequent sections, we focus on queries that satisfy \autoref{eq:model_preference_check}: those the full model $f_\theta$ has learned while the No-$\mathcal{X}$ model $f_{\theta^-}$ has not, i.e., those `attributable' to $\mathcal{X}$.}

\subsection{RQ1: Do IFs surface the training data with the target property \texorpdfstring{$\mathcal{X}$}~?}\label{sec:results_rq1}

\begin{quote}\itshape
Yes in aggregate, but effectiveness varies by phenomenon.
\end{quote}

\paragraph{Aggregated contrast.} \autoref{fig:presence_vs_specifity} shows that at the aggregate level, IFs concentrate on the target-property data more than on the base corpus: the normalized contrast $\Delta\mathcal{I}/\sigma$ is positive and larger over $\mathcal{D}_\mathcal{X}$ than over $\mathcal{D}_{\text{base}}$\footnote{\label{fn:implementation_aggregated_contrast} We implement the condition $\Delta\mathcal{I} \lesssim 0$ for base points as $\Delta\mathcal{I}$ being smaller over $\mathcal{D}_{\text{base}}$ than over $\mathcal{D}_\mathcal{X}$; a fixed threshold would be an alternative but would itself require justification.} for the large majority of phenomenon-method-budget combinations. Hence \autoref{eq:presence_vs_specifity} holds in most cases  (with the exception of NPIs with GradSim method, and binding reflexives with GradSim and K-FAC).

\paragraph{Top-\emph{k} recovery.}

\begin{figure}[!t]
    \centering
    \includegraphics[width=1\linewidth]{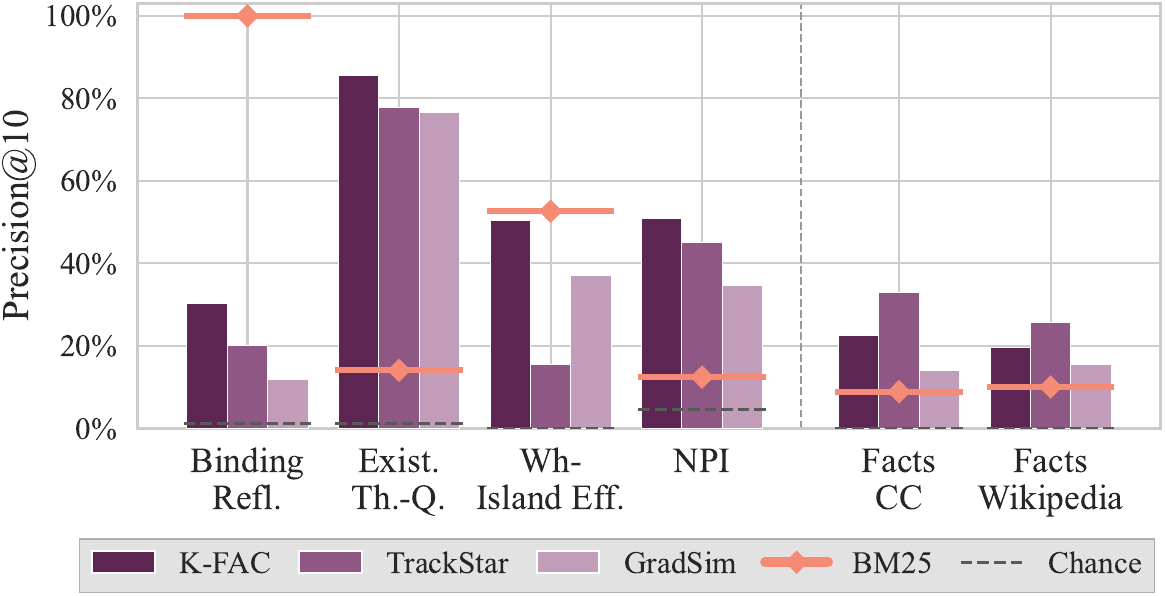}
    \caption{\textbf{Precision@10 of the influence contrast $\Delta \mathcal{I}$} for the 5.6B token Common Corpus model on the four linguistic phenomena and the two factual models. Full results across multiple $k$ and token budgets are available in app.~\autoref{fig:prec_k_full_linguistics} and \autoref{fig:prec_k_full_facts}.}
    \label{fig:prec_at_k_contrast}
\end{figure}

Per-query top-$k$ recovery on the influence contrast is strongly phenomenon-dependent (\autoref{fig:prec_at_k_contrast}). \textit{Existential there quantifiers} are recovered strongly across all approximations up to $86\%$, \textit{wh-island effects} and \textit{NPIs} reach up to $51\%$, and \textit{binding reflexives} only $30\%$. For \textit{binding reflexives}, however, we observe significantly higher precision for lower token budgets, whereas for the other phenomena the differences among token budgets are modest (\autoref{fig:tier-lines}). %
Facts are rather weakly recovered: Prec@$10$ reaches at most $35\%$ on the Common Corpus subset and $26\%$ on Wikipedia.\footnote{Note that for this analysis a factual hit requires co-occurrence and no single-entity occurrence. This is stricter than our filtering criterion, which also removes single-entity occurrences for lower-frequency entities.}%

\paragraph{Comparison to baseline.} BM25 recovers \textit{binding reflexives} perfectly ($100\%$), is on par with the best attribution method for \textit{wh-island effects}, and falls below it in every other case. Most notably, for facts, BM25 contrast reduces to the answer ($Y$) entity (e.g. Paris is in \textit{France} vs. Paris is in \textit{Germany} $\rightarrow$ \textit{France} vs. \textit{Germany}), so single-entity mentions of $Y$ are scored similarly to co-occurrence hits.\footnote{Under \textit{Correct-only} ranking BM25 improves substantially, supporting this explanation (full results app.~\autoref{fig:prec_k_full_facts_good_only}).}

\paragraph{From contrast to a realistic attribution setting.}%
Computing the influence contrast (\autoref{eq:contrast}) requires the minimal pair with the negative example $s^-$, and knowledge of which example is correct. Since this is unavailable outside of minimal pair setup, we additionally investigate the \emph{Correct-only} ranking on $\mathcal{I}(z, s^+)$. We find that for \textit{existential there quantifiers}, \textit{wh-island effects}, and facts, the two rankings align fairly closely: the positive member alone drives the score, so plain top-$k$ already surfaces $\mathcal{X}$. But for \textit{NPIs} and \textit{binding reflexives} the positive example alone is insufficient, so the contrast is essential. Full results are provided in app.  \autoref{fig:contrast_good_bad_linguistics_kfac} and \autoref{fig:contrast_good_bad_facts_kfac}.

\subsection{RQ2: Which types of non‑$\mathcal{X}$ samples are identified as influential?}
\label{sec:rq3}
\begin{quote}\itshape
Most non-$\mathcal{X}$ examples are indirectly relevant to $\mathcal{X}$, and their type and ratio varies by phenomenon and training budget.
\end{quote}

We analyze top-10 examples identified by K-FAC (the most consistent IF method, as will be shown in \cref{sec:results_rq3}) as influential for each linguistic phenomenon. For each we conducted a manual analysis of the likely similarities to samples, which did not directly have the target property $\mathcal{X}$, and could be identified automatically with our corpus annotation: e.g. the NPIs could be supported by only negation or only polarity items. 

\begin{figure}[!t]
\centering
\includegraphics[width=1\linewidth]{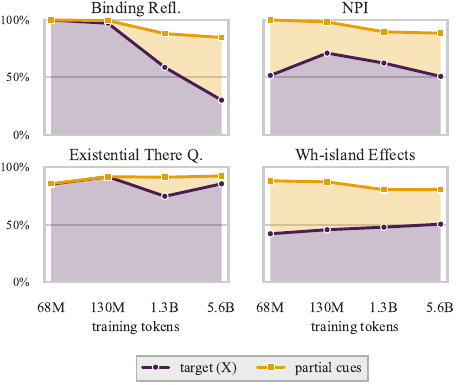}
\caption{\textbf{Share of samples with full target phenomenon $\mathcal{X}$ and partial cues among top-10 retrievals across token
budgets.} These results are for K-FAC method (\cref{sec:results_learned}). %
The plot is cumulative, i.e. partial cue retrievals are stacked on top of target retrievals. The unexplained retrievals are omitted, so the two lines do not sum to 100\%.}
\label{fig:tier-lines}
\end{figure}

\autoref{fig:tier-lines} shows that most of the IF-identified samples that don't explicitly have $\mathcal{X}$ are still linguistically relevant: for example, `\textit{the boy hurt himself}' could be supported by an example like `\textit{the boy hurt his knee}'. The specific supporting constructions and ratios of samples with $\mathcal{X}$ varies by linguistic phenomena and model size.

We further consider, within retrievals from $\mathcal{X}$, whether the specific instances of linguistic phenomena are supported only by themselves or other instances within the same category (e.g. \textit{at all} as a NPI could be supported only by other examples with \textit{at all} or other NPIs such as \textit{anymore}, \textit{ever} etc.). We find the highest ratio of non-identical supporting instances within NPIs ($\sim$30\%), followed by 11.3\% for existential-there, and less than 3\% on average for wh-island effect and binding. Full results are available in app.~\autoref{tab:diversity}.

\paragraph{Facts.}
For facts we manually identify 3 types of possible relations between a training sample and the model output conditioned on a query: co-occurrence of target entities (e.g. a sentence with both France and Paris), only a single entity (e.g. only Paris), and sentences with the target BEAR templates with at most one target entity (e.g. `the capital of [France]'). Using regular expressions, we automatically identify these categories in the top-10 influential training examples for BEAR queries. 

For the models trained on Wikipedia and K-FAC method, in the top-10 influential sentences we find that 66\% contain only single entities relevant to the target fact (e.g. only Paris for <Paris, capital, France> fact). Intuitively, we would expect co-occurrences to be the primary driver of the factual knowledge, but they account for only $19\%$ of top-10 supporting examples. To some extent this is reasonable: single-entity mentions are far more frequent than co-occurrences, supporting the prior-bias learning argument of \citet{christophDataKnowledgeEvaluating2025}. 

The same confound takes a second form: 22\% are nominal sentences without a finite verb, but mostly (-5pp) mentioning a correct entity (e.g., \emph{`Buenos Aires, October 23.'}). We attribute this to the unit-normalisation we apply to IFs, which raises the weight of a single mention in very short sequences. 

Finally, some influential examples map onto the expression of the relation between entities (e.g., \emph{`died in'}, \emph{`capital of'}): either asserting the relation of a different answer entity (\emph{`Germany' + `capital of'}, $4$--$9\%$) or carrying no entity at all (\emph{`capital of'}, $2$--$13\%$). Both indicate the model reliance on texts not directly expressing the target fact, which would be in line with the well-established `hallucination' phenomenon \cite{zhang-etal-2025-sirens}.

\subsection{RQ3: Which IF approximation most reliably identifies relevant training data?}
\label{sec:results_rq3}

\begin{quote}\itshape
K-FAC is the most consistent overall, but tracks TrackStar in the factual setting.
\end{quote}
\autoref{fig:prec_at_k_contrast} summarizes the differences between the approximations. K-FAC performs best on all four linguistic phenomena, TrackStar performs best on facts, and GradSim is consistently the weakest. K-FAC is the most consistent overall: strongest on the linguistic phenomena and close to TrackStar on facts. This aligns with prior work: (E)K-FAC has lower approximation error than the identity (GradSim) \citep{hongBetterHessiansMatter2025}. In fine-tuning setups, it also aligns more closely with the LOO objective than TrackStar \citep{quirkeBergsonOpenSource2026}, though identity and EK-FAC often match \citep{jaburi2025mitigating}. None of these works, however, compares all three methods in pretraining.\footnote{Supplementary results comparing different (E)K-FAC computation approaches are available in \autoref{tab:ekfac_ablation} and \autoref{fig:ekfac_correlation} in the appendix.}

\section{Conclusion}
This study is the first to systematically evaluate whether IFs identify the training examples with properties relevant to the output, based on controlled experiments with filtered training data. Our results suggest that overall IFs do identify the relevant training samples. However, it is a mix of samples with theoretically defined properties and various phenomenon-specific kinds of supporting evidence, and the ratio of that varies by phenomenon and model size. Of the three IF implementations we tested, K-FAC is overall the most consistent  in identifying the training data with the target properties, but there is room for improvement in future work on developing better approximations for IFs.

\section*{Limitations}
\label{sec:limitations}
\paragraph{Scale and generalizability.} We train on small models and limited data (68M--5.6B tokens) using a single (GPT-2) architecture, so it is unclear whether the same patterns hold at larger scale or across architectures. However, prior work suggests that IFs surface more abstract structures at larger scale \citep{grosseStudyingLargeLanguage2023, ruisProceduralKnowledgePretraining2025a}, aligning with our findings: e.g. structures like \textit{binding reflexives} surface more related structures with scale (\cref{sec:rq3}), while pattern-reducible ones like \textit{existential there quantifiers} should stay more consistent. We use GPT-2 because it is common in the literature on learning acquisition \citep{warstadtBLiMPBenchmarkLinguistic2023, christophDataKnowledgeEvaluating2025}. 

Generalizability across data is likewise not guaranteed, and it would be worth investigating how well the results transfer to other datasets. 
More fundamentally, we deliberately select phenomena that plausibly depend on specific training examples, yet even for these, interactions across examples play a considerable role. For facts in particular, prior work and ours show that isolating the property is hard. 
Future work should investigate the nature of influential examples for phenomena more prone to being learned through such interactions, than the ones in focus of this study.

\paragraph{Coverage.} At several steps we cannot guarantee full coverage of the relevant data space. Linguistic filtering misses items due to the nature of syntactic filters (app.~\ref{subsec:appendix_linguistic}), and in the factual setting entity occurrences take diverse surface forms that our alias- and demonym-based matching may not fully capture (app.~\ref{subsec:appendix_factual}). Attribution also runs over a subsample of the base corpus, so we may miss base data that would change the results.

\paragraph{Scope of attribution methods.} This study focuses on three of the most prominent Hessian approximations for IFs, GradSim, TrackStar, and K-FAC. We do not consider other methods, particularly those discussed in footnote \ref{fn:unrolling}. We also use the standard influence formula. Future work could investigate whether extensions to it, such as those proposed by \citet{kowalConceptInfluenceLeveraging2026} or \citet{heoInteractionAwareInfluenceFunctions2026}, would yield different results. Finally, we do not sweep hyperparameters for the methods investigated, except for the (E)K-FAC ablation (details available in the supplementary materials: \autoref{tab:ekfac_ablation} and \autoref{fig:ekfac_correlation}).

\paragraph{Training stochasticity.} Due to computational constraints, we train and attribute on a single seed, but IFs are known to be sensitive to training stochasticity \citep{mlodozeniecDistributionalTrainingData2025}. We partly account for this by retraining
from scratch for each token budget rather than branching from shared checkpoints, but this does not fully cover what a complete picture would require. 

\paragraph{Training order and ratio.} This study does not consider when a given training example was observed during training. Future work should investigate whether the ability of IFs to identify qualitatively relevant training examples varies depending on whether the relevant examples were observed early or late in the training, and how many examples of a given type were generally available to the model during the training. 
\section*{Ethical Considerations}
\label{sec:ethics}
\paragraph{Data sourcing.} Our models are trained on an English
subset of Common Corpus \citep{langlaisCommonCorpusLargest2026}\footnote{\url{https://huggingface.co/datasets/PleIAs/common_corpus}} and an English
Wikipedia dump \citep{wikimedia_english_wikipedia_dump_2023_11_01}.\footnote{\url{https://huggingface.co/datasets/wikimedia/wikipedia}} The Wikipedia dump is licensed under CC-BY-SA. Common Corpus consists exclusively of public domain and permissively licensed text, for which our sample will retain the original licenses. We refer to the original Common Corpus work by \citet{langlaisCommonCorpusLargest2026} for efforts to remove personally identifiable information and toxic content (sec. 4.4, 4.5).

\paragraph{Potential risks.}

The main ethical risk arising from this work concerns the interpretation and downstream use of influence estimates. Influence functions may support beneficial applications such as model debugging, data curation, unlearning, and the attribution of model outputs for creator compensation. However, identifying data associated with particular model behaviours could also facilitate targeted data manipulation or strategically harmful removal requests.

Moreover, given the limited faithfulness of influence functions indicated by our results and prior work, even well-intentioned applications carry risks. In particular, influence functions may misidentify the data responsible for a model’s behaviour, potentially resulting in unfair compensation, unwarranted blame, or the removal of data incorrectly classified as irrelevant or harmful.

\paragraph{Environmental costs.} We used \textsc{CodeCarbon} \citep{benoit_courty_2024_11171501} to estimate the environmental cost of our experiments, run on NVIDIA H100 NVL and L40S GPUs. Across measured runs, consumption averaged 0.62\,kWh and 95\,g CO\textsubscript{2}eq per GPU-hour. Our experiments required 1{,}730 GPU-hours in total, including failed runs and resubmissions, amounting to approximately 1{,}080\,kWh and 160\,kg CO\textsubscript{2}eq. Reproducing all reported experiments would take approximately 730 GPU-hours, or roughly 460\,kWh and 70\,kg CO\textsubscript{2}eq. 
\section*{Acknowledgments}

We thank Louis Jaburi for general advice and helpful comments throughout the project, Bertram Højer for title inspiration, and Max Müller-Eberstein for help with filtering the Common Corpus. The research was supported by a research grant (VIL60860) from Villum Fonden.

\paragraph{AI Assistance.} We used Anthropic's Claude models as assistive tools during this project. Claude Opus 4.8 was used to refine the grammar and style of early drafts, which were subsequently fully revised by the authors. The research questions, experimental design, analysis, and interpretation of results are entirely their own. Claude Sonnet 5 and partly Claude Opus 4.8 assisted with implementation, primarily boilerplate around external libraries such as \textsc{Bergson} \citep{quirkeBergsonOpenSource2026}, plotting, and refactoring. All experimental logic was specified by the authors, and all AI-assisted output was reviewed and validated by them. The authors take full responsibility for the content of this work.

\bibliography{custom}

\appendix
\crefalias{section}{appsec}
\section{IFs and Hessian Approximations: Derivations}
\label{sec:appendix_methods}

This section gives a comprehensive derivation of the influence
function from its underlying leave-one-out (LOO) objective, and then provides a detailed overview of the three curvature approximations utilized in this work.

\subsection{From Leave-One-Out to IFs}
\label{subsec:appendix_if}

A natural way to approach data attribution is the \emph{leave-one-out} (LOO)
principle: how does a differentiable observable of the model's parameters $\phi(\theta)$ from \cref{sec:background}
change when a single training point $z$ is removed from
$\mathcal{D}_\mathrm{train}$ and the model is retrained from scratch? In most of the recent influence work, the observable $\phi(\theta) = \mathcal{L}(\theta, z')$ is the query loss.

Writing the trained parameters as the empirical risk minimiser
\begin{equation}
    \theta^* = \arg\min_\theta \frac{1}{n} \sum_{i=1}^{n}
        \mathcal{L}(\theta, z_i) ,
\end{equation}
and the LOO-retrained parameters as
\begin{equation}
    \label{eq:loo_param}
    \theta^*_{-z} = \arg\min_\theta \frac{1}{n}
        \sum_{z_i \neq z} \mathcal{L}(\theta, z_i) ,
\end{equation}
the LOO objective is the resulting change in the observable,
\begin{equation}
    \label{eq:loo}
    \mathcal{I}_{\mathrm{LOO}}(z, z') =
        \phi(\theta^*_{-z}) - \phi(\theta^*) .
\end{equation}
With $\phi = \mathcal{L}(\cdot, z')$ this is the increase in query loss upon
removal: a positive value means that removing $z$ raises the query loss, i.e.,
$z$ supported the model's behaviour on $z'$.

Evaluating \autoref{eq:loo} for every $z$ requires $n$ separate retrainings,
which is intractable at LLM scale.
IFs \citep{kohUnderstandingBlackboxPredictions2017} approximate
$\mathcal{I}_{\mathrm{LOO}}$ without retraining, instead studying a continuous up- / downweighting of $z$ through the response function
\begin{equation}
    \label{eq:response}
    \theta^*_\epsilon = \arg\min_\theta \frac{1}{n} \sum_{i=1}^{n}
        \mathcal{L}(\theta, z_i) + \epsilon\, \mathcal{L}(\theta, z) ,
\end{equation}
where $\epsilon = 0$ recovers $\theta^*$ and $\epsilon = -\tfrac{1}{n}$ corresponds to removing $z$ entirely, matching the LOO setup.
The minimiser of \autoref{eq:response} satisfies the first-order optimality
condition
\begin{equation}
    \frac{1}{n} \sum_{i=1}^{n} \nabla_\theta \mathcal{L}(\theta^*_\epsilon, z_i)
        + \epsilon\, \nabla_\theta \mathcal{L}(\theta^*_\epsilon, z) = 0 .
\end{equation}
Differentiating this condition with respect to $\epsilon$ at $\epsilon = 0$,
applying the implicit function theorem to $\theta^*_\epsilon$, and assuming
$\theta^*$ is a true (global) minimum of the training loss, gives
\begin{equation}
    H_{\theta^*}\, \frac{d\theta^*_\epsilon}{d\epsilon}\Big|_{\epsilon=0}
        + \nabla_\theta \mathcal{L}(\theta^*, z) = 0 ,
\end{equation}
with $H_{\theta^*} = \frac{1}{n} \sum_i \nabla^2_\theta
\mathcal{L}(\theta^*, z_i)$ the training-loss Hessian. Solving for the
parameter sensitivity yields
\begin{equation}
    \label{eq:param_sensitivity}
    \frac{d\theta^*_\epsilon}{d\epsilon}\Big|_{\epsilon=0}
        = -H_{\theta^*}^{-1}\, \nabla_\theta \mathcal{L}(\theta^*, z) .
\end{equation}
The chain rule then propagates this shift to the observable,
\begin{equation}
    \label{eq:phi_sensitivity}
    \frac{d\,\phi(\theta^*_\epsilon)}{d\epsilon}\Big|_{\epsilon=0}
        = -\, \nabla_\theta \phi(\theta^*)^\top H_{\theta^*}^{-1}\,
        \nabla_\theta \mathcal{L}(\theta^*, z) .
\end{equation}
In the standard setting $\phi(\theta) = \mathcal{L}(\theta, z')$, so
$\nabla_\theta \phi(\theta^*) = \nabla_\theta \mathcal{L}(\theta^*, z')$ is the
query gradient.

Since $\theta^*_{-z} = \theta^*_\epsilon$ at $\epsilon = -\tfrac{1}{n}$, a
first-order expansion of \autoref{eq:loo} around $\epsilon = 0$ then gives
\begin{align}
    \mathcal{I}_{\mathrm{LOO}}(z, z')
        &\approx -\tfrac{1}{n}
        \frac{d\,\phi(\theta^*_\epsilon)}{d\epsilon}\Big|_{\epsilon=0}
        \nonumber \\
        &= \tfrac{1}{n}\, \nabla_\theta \phi(\theta^*)^\top H_{\theta^*}^{-1}\,
        \nabla_\theta \mathcal{L}(\theta^*, z)
        \nonumber
\end{align}
The standard setting $\phi(\theta) = \mathcal{L}(\theta, z')$ yields then
\begin{align}
    \mathcal{I}_{\mathrm{LOO}}(z, z') = \tfrac{1}{n}\,
        \nabla_\theta \mathcal{L}(\theta^*, z')^\top H_{\theta^*}^{-1}\,
        \nabla_\theta \mathcal{L}(\theta^*, z) .
\end{align}
Removing the constant factor $\frac{1}{n}$ yields the influence score in \autoref{eq:if}.

\subsection{Cosine Similarity \& Random Projections}
\label{subsec:appendix_cosine_projections}

Writing the parameter gradient as $g_z = \nabla_\theta \mathcal{L}(\theta^*, z)
\in \mathbb{R}^p$, the influence score of \autoref{eq:if} reads
$\mathcal{I}(z, z') = g_{z'}^\top H_{\theta^*}^{-1} g_z$. Whenever
$H_{\theta^*}$ is symmetric positive definite it has a unique symmetric square
root, so the score is a plain inner product of \emph{whitened} gradients
$\tilde{g}_{z} = H_{\theta^*}^{-1/2} g_z$,
\begin{equation}
    \label{eq:if_whitened}
    \mathcal{I}(z, z') = \tilde{g}_{z'}^\top \tilde{g}_{z} .
\end{equation}
Following \citet{changScalableInfluenceFact2024}, we additionally
unit-normalise these gradients, which gives the quantity we report throughout,
\begin{equation}
    \label{eq:if_cosine}
    \mathcal{I}(z, z') \approx
    \cos\!\left(\tilde{g}_{z'}, \tilde{g}_{z}\right) .
\end{equation}

\autoref{eq:if_whitened} and \autoref{eq:if_cosine} are defined over the full
parameter space, with $p$ on the order of $10^8$: storing one gradient per
training point does not scale, and forming or inverting the $p \times p$
Hessian is infeasible. Following \citet{parkTRAKAttributingModel2023}, we
therefore compress gradients by random projections $\Phi \in \mathbb{R}^{k
\times p}$, $k \ll p$. By the Johnson-Lindenstrauss lemma
\citep{johnsonExtensionsLipschitzMappings1984}, inner products are
approximately preserved in expectation, so it suffices to store the sketch
$\Phi g_z \in \mathbb{R}^k$ once per training point and reuse it for every
query.

A dense $\Phi$ would itself be too large to store. Following
\citet{changScalableInfluenceFact2024}, we apply $\Phi$ as a block-diagonal
structure with one block per weight module, and factor each block into two
per-side matrices $\Phi_L, \Phi_R$, so that projecting a block gradient $G_W$
costs only two small matrix products,
\begin{equation}
    \label{eq:two_sided_proj}
    \bar{G}_W = \Phi_L\, G_W\, \Phi_R^\top ,
\end{equation}
rather than one dense $k \times p$ multiply. The entries of $\Phi_L, \Phi_R$
are drawn i.i.d.\ from $\{-1, +1\}$, which preserves inner products with the
same guarantees as a Gaussian sketch
\citep{achlioptasDatabasefriendlyRandomProjections2003}. We write $\hat{g}_z
\in \mathbb{R}^k$ for the projected gradient obtained by flattening and
concatenating the sketched blocks.

Random projection also makes curvature estimation cheaper. Rather than forming
$H$ in parameter space and then projecting it, we work directly with a
\emph{compressed} pseudo-Hessian $\hat{H} \in \mathbb{R}^{k \times k}$
estimated in the projected space \citep{changScalableInfluenceFact2024}, again
block-diagonal so that $\hat{H}^{-1/2}$ stays cheap. \autoref{eq:if_cosine}
becomes
\begin{equation}
    \label{eq:if_compressed}
    \mathcal{I}(z, z') \approx
    \cos\!\left(
        \hat{H}^{-1/2} \hat{g}_{z'},\;
        \hat{H}^{-1/2} \hat{g}_{z}
    \right) .
\end{equation}
The three methods below differ only in how they instantiate $\hat{H}$.

\subsection{Hessian Approximations}
\label{subsec:appendix_curvature}

\begin{figure*}[!h]
    \centering
    \includegraphics[width=1\linewidth]{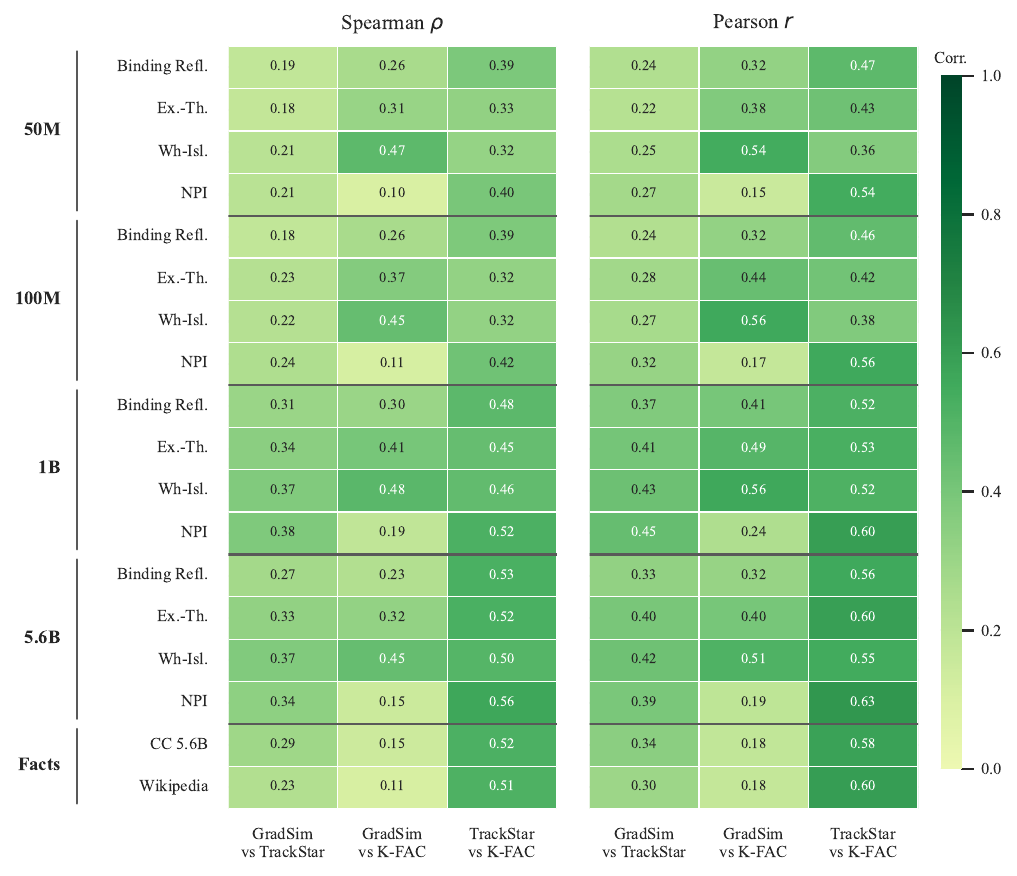}
    \caption{\textbf{Agreement of GradSim, TrackStar, and K-FAC.} Per-query-pair Spearman
    $\rho$ (left) and Pearson $r$ (right) on \textit{all queries} between the $\Delta\mathcal{I}$ that two methods assign to the candidate sentences in $\mathcal{D}_\mathcal{X}$, computed independently for each query pair and averaged over pairs within a panel. \emph{K-FAC} is the fused scheme~(iii) variant used in the main results.}
   \label{fig:method_correlation}
\end{figure*}

We compare three choices of $\hat{H}$, in increasing order of how faithfully
they model the loss curvature. Each is plugged into
\autoref{eq:if_compressed}; they differ in what they fit and at what cost. \autoref{fig:method_correlation} shows the correlation between the contrastive influence scores at the individual query–training-point level.

\subsubsection*{Gradient Similarity (GradSim)}

The simplest choice makes no curvature correction at all, $\hat{H} = I$, so
that \autoref{eq:if_compressed} reduces to the cosine similarity of the raw
projected gradients, $\mathcal{I}(z, z') \approx \cos(\hat{g}_{z'},
\hat{g}_{z})$.

This estimator follows \citet{pruthiEstimatingTrainingData2020}, who derive an
influence estimator by tracing the training trajectory: a first-order expansion
of the SGD update $w_{t+1} - w_t = -\eta_t g_{z,t}$ gives a training point's
contribution to the query loss over training as a learning-rate-weighted sum of
gradient dot products across checkpoints,
\begin{equation}
    \mathrm{TracIn}(z, z') = \sum_{t:\, z_t = z} \eta_t\,
        g_{z',t}^\top g_{z,t} ,
\end{equation}
with $g_{z,t} = \nabla_\theta \mathcal{L}(w_t, z)$. We evaluate gradients only
at the final checkpoint $\theta^*$ and assume constant step size for simplicity, which recovers $\hat{H} = I$. This sounds like a naive approach, but there are reasons to believe that it is not as poor as a choice as it seems \citep{grunefeldIsotropicApproachEfficient2026}.

GradSim requires no fitting pass
at all: its only cost is one projected gradient per point plus the scoring dot
products, making it by far the cheapest method.

\subsubsection*{TrackStar}

TrackStar \citep{changScalableInfluenceFact2024} estimates $\hat{H}$ directly
in the projected space, as the autocorrelation (second moment) of the
compressed gradients. Over the $n$ training points and the $2m$ query
sentences (each of the $m$ minimal pairs contributes a positive and a negative
sentence) it fits
\begin{align}
    \hat{H}_{\mathrm{train}} &= \frac{1}{n}\sum_{i=1}^{n}
        \hat{g}_{z_i}\, \hat{g}_{z_i}^\top , \\
    \hat{H}_{\mathrm{query}} &= \frac{1}{2m}\sum_{j=1}^{2m}
        \hat{g}_{z'_j}\, \hat{g}_{z'_j}^\top .
\end{align}
Because these gradients are taken at the \emph{observed} training labels rather
than at labels sampled from the model's predictive distribution, $\hat{H}$ is a
(compressed) \emph{empirical} Fisher, which in general differs from the true
Fisher and only loosely tracks the loss Hessian near the optimum
\citep{kunstnerLimitationsEmpiricalFisher2020}; (E)K-FAC (below) instead targets
the true Fisher.

To suppress gradient directions that would otherwise dominate the score,
TrackStar mixes the two estimates, $\hat{H}_{\mathrm{mix}} = \alpha
\hat{H}_{\mathrm{query}} + (1-\alpha) \hat{H}_{\mathrm{train}}$, with the query
weight $\alpha$ selected automatically so that the 1,000 most query-dominant
components are downweighted \citep{changScalableInfluenceFact2024}.
$\hat{H}_{\mathrm{mix}}^{-1/2}$ is applied to both sides of
\autoref{eq:if_compressed}. The original formulation also divides gradients by
the optimiser's second-moment (Adafactor) estimate; we omit this term,
following \citet{quirkeBergsonOpenSource2026}. Fitting $\hat{H}$ costs a single pass over the
training gradients, and the block-diagonal structure keeps the inverse square
root cheap.

\subsubsection*{K-FAC}

K-FAC \citep{martensOptimizingNeuralNetworks2015} models the curvature most
faithfully in theory, at the cost of more compute: unlike the two methods
above, it approximates the Hessian itself.

Because the loss Hessian of a neural network is neither guaranteed positive
definite nor tractable \citep{sagunEmpiricalAnalysisHessian2018,
baeIfInfluenceFunctions2022}, K-FAC replaces it with the Gauss-Newton matrix,
which for the cross-entropy loss equals the Fisher information matrix,
\begin{equation}
    F(\theta) = \mathbb{E}_{y \sim p_\theta}\!\left[
        \nabla_\theta \log p_\theta(y)\,
        \nabla_\theta \log p_\theta(y)^\top
    \right] .
\end{equation}
The expectation is over labels drawn from the model's own predictive
distribution (the \emph{true} Fisher), in contrast to TrackStar's empirical
Fisher; consequently $F$ cannot simply be estimated as an outer product of
compressed gradients. $F$ is positive semidefinite by construction, and a
damping term $\lambda > 0$ makes it strictly positive definite. 

K-FAC then makes $F$ tractable in two steps. First, it assumes a block-diagonal
structure across layers, one block $F_i$ per weight matrix $W_i \in
\mathbb{R}^{d_\mathrm{in} \times d_\mathrm{out}}$. 

Second, it exploits the fact that the per-example gradient of $W_i$ has the outer-product form $\nabla_{W_i}\mathcal{L} = a_{i-1} Ds_i^\top$, the layer input $a_{i-1}$ times the pre-activation pseudo-gradient $Ds_i$.\footnote{For a linear layer with
pre-activations $s_i = W_i^\top a_{i-1}$, the input $a_{i-1} \in
\mathbb{R}^{d_\mathrm{in}}$ is the activation produced by the preceding layer,
and $Ds_i = \nabla_{s_i}\mathcal{L} \in \mathbb{R}^{d_\mathrm{out}}$ is the
gradient of the loss with respect to those pre-activations, i.e., precisely the
quantity backpropagation already computes at that layer, so both factors are
available without extra cost. The chain rule then gives $\nabla_{W_i}\mathcal{L}
= a_{i-1} Ds_i^\top$. We follow \citet{martensOptimizingNeuralNetworks2015} in
calling $Ds_i$ a \emph{pseudo}-gradient: for the true Fisher, $\mathcal{L}$ is
evaluated at labels sampled from $p_\theta$ rather than at the observed ones, so
it is not the gradient of the actual training loss.} Assuming these are statistically
independent, each block factorises into a Kronecker product,
\begin{align}
    F_i &\approx A_i \otimes S_i, \label{eq:kfac_factor} \\
    A_i &= \mathbb{E}[a_{i-1} a_{i-1}^\top],\quad
    S_i  = \mathbb{E}[Ds_i\, Ds_i^\top] .
\end{align}
The factors $A_i$ ($d_\mathrm{in} \times d_\mathrm{in}$) and $S_i$
($d_\mathrm{out} \times d_\mathrm{out}$) are far smaller than $F_i$ and can be
estimated in a single pass over the training data. We fit them for every linear
layer except the language-model head (the token embedding is likewise
excluded).

Eigendecomposing the factors ($A_i = Q_A \Lambda_A Q_A^\top$, $S_i = Q_S \Lambda_S Q_S^\top$), the Kronecker structure gives a closed-form
inverse-Hessian-vector product per block,
\begin{multline}
    \label{eq:kfac_ihvp}
 (F_i + \lambda I)^{-1}\,\mathrm{vec}(G_{z,i}) \\
    = \mathrm{vec}\!\left(
        Q_A \bigl(Q_A^\top G_{z,i}\, Q_S \odot \Lambda^{-1}\bigr) Q_S^\top
    \right),
\end{multline}
where $G_{z,i}$ is the gradient of block $i$, $\odot$ is the element-wise
product, and $\Lambda^{-1}$ element-wise inverts $\Lambda_A \otimes \Lambda_S$
after adding $\lambda$. This never forms or inverts the full block. Stacking
the blocks yields the approximation to $H^{-1}g_z$ used in \autoref{eq:if}.

\paragraph{EK-FAC.}
\label{sec:appendix_lambda}
The independence assumption behind \autoref{eq:kfac_factor} makes $\Lambda_A
\otimes \Lambda_S$ only an approximation to the true eigenvalues of $F_i$ in
the Kronecker eigenbasis. EK-FAC \citep{georgeFastApproximateNatural2018}
corrects this by re-estimating the diagonal directly in that eigenbasis, at the
cost of one extra pass over the data.

\paragraph{(E)K-FAC with Whitening and Random Projections.}
Unlike GradSim and TrackStar, which are defined directly on the compressed
gradients of app.~\ref{subsec:appendix_cosine_projections}, (E)K-FAC was formulated in
full parameter space and without the whitening of
app.~\ref{subsec:appendix_cosine_projections} \citep{martensOptimizingNeuralNetworks2015,
georgeFastApproximateNatural2018}: the factors $A_i, S_i$ live in the layer
dimensions $d_\mathrm{in}, d_\mathrm{out}$, and $H^{-1}$ is applied to
full-dimensional gradients.
 
Whitening is cheap to add, since the Kronecker structure carries over to the
symmetric square root,
\begin{equation}
    \label{eq:kron_sqrt}
    F_i^{-1/2} \approx (A_i \otimes S_i)^{-1/2}
        = A_i^{-1/2} \otimes S_i^{-1/2} ,
\end{equation}
with $A_i^{-1/2} = Q_A \Lambda_A^{-1/2} Q_A^\top$ and $S_i^{-1/2} = Q_S
\Lambda_S^{-1/2} Q_S^\top$; concretely, this amounts to replacing
$\Lambda^{-1}$ by $\Lambda^{-1/2}$ in \autoref{eq:kfac_ihvp} (where $\Lambda$
already carries the damping $\lambda$) on both the query and the training side,
so the full K-FAC machinery carries over unchanged. EK-FAC's corrected
eigenvalues no longer factorise across the two sides as in
\autoref{eq:kron_sqrt}, but they remain diagonal in the Kronecker eigenbasis,
so the square root is still available.
 
Scale remains a problem: a full-dimensional gradient must be materialised for every training point at scoring time and cannot be cached for reuse across
queries. Two remedies are available: Two remedies are available: amortising one pass over many queries, or storing a compressed sketch per point. This gives the three schemes
below, in decreasing order of cost.

\emph{(i) Query batching.} \citet{grosseStudyingLargeLanguage2023} propose batching the queries so that a single pass over the training data scores a whole query batch at once, sharing the training-gradient computation across the batch. To fit a batch in memory they approximate each preconditioned query gradient as low-rank. They report this to incur only small error, but like the following two schemes it is not approximation-free.  \textit{Cost:} The Kronecker factors of the `Hessian' are fitted in full parameter space, and scoring dominates: the low-rank matrix-vector products are repeated once per query batch, so time and memory both grow with the number of queries.

\begin{figure*}[!h]
    \centering
    \includegraphics[width=1\linewidth]{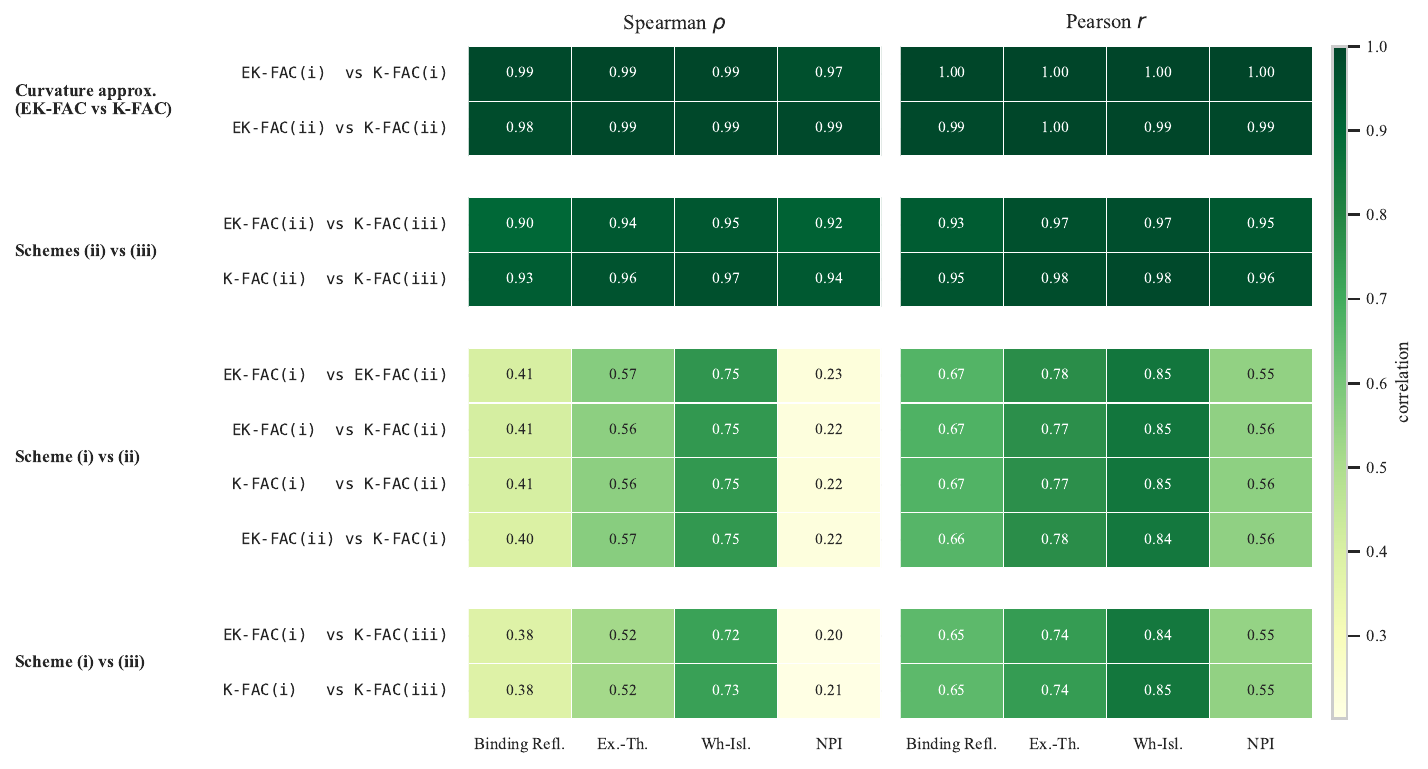}
    \caption{\textbf{Agreement between the three (E)K-FAC application schemes
    described in Appendix~\ref{subsec:appendix_curvature}.} Per-query Spearman
    $\rho$ (left) and Pearson $r$ (right) between the $\Delta\mathcal{I}$ scores
    that two schemes assign to the candidate sentences in
    $\mathcal{D}_\mathcal{X}$, averaged over queries. Changing the curvature
    (EK-FAC vs.\ K-FAC) or moving from the sketched to the fused scheme leaves
    the ranking largely intact. Query batching, however, shows notable gaps. To
    limit compute, we scored 16 query pairs at $\lambda = 0.1$ on the 5.6B
    model, with 500{,}000 candidate sentences per phenomenon.}
    \label{fig:ekfac_scheme_agreement}
   \label{fig:ekfac_correlation}
\end{figure*}

\emph{(ii) Sketched application (whiten, then project).}
Random projections can be introduced \emph{post-hoc}: whiten each block exactly
as in scheme (i), stack the results into the full whitened gradient
$\tilde{g}_z$, and only then reduce its dimension for storage and scoring,
$\bar{g}_z = \Phi\,\tilde{g}_z$. The Kronecker factors are still fitted and
applied in full parameter space. Because whitening happens before any
information is lost to the projection, this scheme stays compatible with the
EK-FAC eigenvalue correction. \textit{Cost:} Fitting is as expensive as in scheme (i), but scoring is amortised: each training point is reduced once to a $k$-dimensional sketch and reused for every query, leaving only $k$-dimensional dot products. That single pass is still costly, since the full whitened gradient is formed before projecting.

\emph{(iii) Fused application (project and whiten in one step).}
We can avoid whitening in full parameter space entirely by folding it into the
projection, at the price of a looser approximation. By
\autoref{eq:kron_sqrt}, the two-sided projection $\Phi_L, \Phi_R$ and the
two-sided whitening $A_i^{-1/2}, S_i^{-1/2}$ act on the same two sides, so each
side can be precombined \emph{once}, after the factors are fitted, into
$M_L^{(i)} = \Phi_L^{(i)} A_i^{-1/2}$ and $M_R^{(i)} = \Phi_R^{(i)}
S_i^{-1/2}$. Each block is then whitened and sketched in a single pair of
matmuls,
\begin{equation}
    \bar{G}_{z,i} = M_L^{(i)}\, G_{z,i}\, M_R^{(i)\top} ,
\end{equation}
the projected counterpart of the whitened block in scheme (ii), but obtained
without ever forming the full-dimensional whitened gradient. \textit{Cost:} Fitting costs the same as schemes (i) and (ii), since the same factors are estimated; the saving is entirely at scoring, where each block collapses to small matmuls.

Damping, however, can no longer be shared between the two factors. Schemes (i)
and (ii) add $\lambda$ to the \emph{combined} eigenvalues $\Lambda_A \otimes
\Lambda_S$ (\autoref{eq:kfac_ihvp}). However, for scheme (iii) the per-side split of \autoref{eq:kron_sqrt} has no access to this joint spectrum, so we instead damp each factor on its own,
\begin{equation}
    A_i^{-1/2} = Q_A\bigl(\Lambda_A + \lambda\,\bar{\Lambda}_A I\bigr)^{-1/2}
        Q_A^\top ,
\end{equation}
and likewise for $S_i^{-1/2}$, where $\bar{\Lambda}_A$ is the mean eigenvalue
of $A_i$. Finally, since EK-FAC replaces $\Lambda_A \otimes \Lambda_S$ with a
corrected diagonal that no longer factorises across the two sides, it cannot be absorbed into $M_L, M_R$; scheme (iii) is therefore restricted to plain K-FAC.

\paragraph{Agreement.} \autoref{fig:ekfac_scheme_agreement} compares all three schemes on a small set of 16 query pairs ranked on 500{,}000 sentences sampled from the same
distribution as in our main experiments, using a rank-32 low-rank approximation for the batched query gradients in scheme (i). Note that 16 queries per phenomenon is a small set, even at 500{,}000 candidates each. We therefore read it as a proxy for how the schemes relate rather than a ranking of them.

EK-FAC and K-FAC agree at $\rho \geq 0.97$ within a scheme. Schemes (ii) and
(iii) agree nearly as closely ($\rho = 0.90$--$0.97$), while scheme (i) shows
notable gaps to the others ($\rho = 0.20$--$0.75$, $r = 0.55$--$0.85$). As a
sanity check that query batching does not overturn our main findings, we
computed Precision@10 on the influence contrast for these 16 queries: it is
never best on any phenomenon and trails the best scheme by at most 6\,pp, so it
offers no gain.

For the sketched approaches we run a more thorough analysis: \autoref{tab:ekfac_ablation} reports Precision@10 for schemes (ii) and (iii) over all existing query pairs per phenomenon and two values of $\lambda$. We leave query batching out here, since its cost grows with the number of queries and does not scale to this setting.

\newcolumntype{L}{>{\raggedright\arraybackslash}X}

\begin{table}[!t]
\centering
\scriptsize
\setlength{\tabcolsep}{3.5pt}
\begin{tabularx}{\columnwidth}{@{}L d{1.2} cccc|c@{}}
\toprule
Method & \multicolumn{1}{c}{$\lambda$}
       & Bind. & Ex.-Th. & Wh-Isl. & NPI & Mean \\
\midrule
\multirow{2}{*}{EK-FAC (sketched)}
  & 0.01 & 35.3 & 88.8 & \textbf{48.5} & 48.8 & 55.4 \\
  & 0.1  & 37.3 & \textbf{90.0} & 48.1 & 48.8 & \textbf{56.1} \\
\midrule
\multirow{2}{*}{K-FAC (sketched)}
  & 0.01 & 35.7 & 86.8 & 47.6 & \textbf{50.0} & 55.0 \\
  & 0.1  & \textbf{37.6} & 88.8 & 48.1 & 49.1 & 55.9 \\
\midrule
\midrule
\multirow{2}{*}{K-FAC (fused)}
  & 0.01 & 32.2 & 82.9 & 47.2 & 48.5 & 52.7 \\
  & 0.1  & 33.6 & 86.5 & 46.8 & 49.4 & 54.1 \\
\bottomrule
\end{tabularx}
\caption{\textbf{Retrieval quality of the sketched (ii) and fused (iii) (E)K-FAC application schemes described in Appendix~\ref{subsec:appendix_curvature}. EK-FAC is marginally better than the uncorrected versions, but gaps are small and do not repay its extra Hessian cost.} Precision@10 (\%) of the influence contrast $\Delta\mathcal{I}$ across the four linguistic phenomena, per scheme and damping $\lambda$, on the 5.6B Common Corpus subset model. As in most parts of this paper, Precision@10 is computed on the \emph{Full correct, No-$\mathcal{X}$ wrong} query pairs only, those `attributable' to $\mathcal{X}$. We fit and score on a reduced sample of 500{,}000 sentences per phenomenon, rather than the 5{,}000{,}000 used in the main experiments, keeping the $\mathcal{D}_\mathcal{X}$-to-$\mathcal{D}_{base}$ proportions of that setup; this yields results close to the main experiments (max.\ 4\,pp of disagreement for the default $\lambda = 0.1$, fused K-FAC). Scheme (iii) is K-FAC-only by construction. Query batching (i) is omitted here: it does not scale well for multi-query setting, and is compared separately in \autoref{fig:ekfac_scheme_agreement}.}
\label{tab:ekfac_ablation}
\end{table}

\section{Scoring Metric and Learned Verdict}
\label{appendix:metric_verdict}

\paragraph{Scoring metric.}
We score a sequence by its summed, unnormalized log-probability $\log p_f(s)$. A model `prefers' a sequence when it assigns it a higher log-probability than its counterpart (see \autoref{eq:model_preference_check}). This aligns with the
primary metric of the benchmarks we adopt, BLiMP
\citep{warstadtBLiMPBenchmarkLinguistic2023} and BEAR
\citep{wilandBEARUnifiedFramework2024}.

Notably, as also discussed by \citet{wilandBEARUnifiedFramework2024}, this
choice introduces a bias: the summed log-probability favours shorter members
of a pair by design. The natural alternative, length-normalizing the scores,
does not remove the bias but shifts it to a tokenization-dependent one.
In our linguistic setting, for instance, our Goldfish tokenizer
splits \emph{only} into two tokens (\emph{on-ly}) while \emph{even}, remains a single token. So a normalized comparison would shift the preference toward \emph{even} for length-related rather than grammatical reasons. The factual setting has the same issue by construction: entities differ in length (e.g.\ \emph{Stockholm} vs.\ \emph{Paris}) and are therefore split into differing numbers of tokens.

\citet{warstadtBLiMPBenchmarkLinguistic2023} further propose two alternative
strategies, one-prefix and two-prefix, which compare probabilities only at
the point where a minimal pair diverges. Since these still depend on the
tokenization of the differing parts, the concerns above persist. One-prefix and two-prefix scoring is also applicable in every setting (see \citet{warstadtBLiMPBenchmarkLinguistic2023}), and thus offer no clear advantage.

\begin{table*}[!h]
\centering
\small
\setlength{\tabcolsep}{5pt}
\begin{tabular*}{\textwidth}{
  @{\extracolsep{\fill}}
  l l r
  w{r}{1cm} w{r}{1cm} w{r}{1cm}
  w{r}{1cm} w{r}{1cm} w{r}{1cm}
  @{}
}
\toprule
 & & & \multicolumn{3}{c}{\textbf{Full}} & \multicolumn{3}{c}{\textbf{No-$\mathcal{X}$} (filtered)} \\
\cmidrule(lr){4-6}\cmidrule(lr){7-9}
Phenomenon & Tokens & $N$ & FS-LL & LN-LL & $\Delta$ & FS-LL & LN-LL & $\Delta$ \\
\midrule
\midrule
\multicolumn{9}{l}{\textbf{\textit{Linguistic phenomena}}}  \\
\midrule
\multirow{4}{*}{Binding Reflexives}
 & 68M & 1000 & 58.1 & 54.1 & $+4.0$ & 61.7 & 44.6 & $+17.1$ \\
 & 130M & 1000 & 59.6 & 56.7 & $+2.9$ & 56.9 & 44.8 & $+12.1$ \\
 & 1.3B & 1000 & 74.7 & 62.5 & $+12.2$ & 30.5 & 30.1 & $+0.4$ \\
 & 5.6B & 1000 & 75.0 & 62.1 & $+12.9$ & 49.9 & 49.6 & $+0.3$ \\
\midrule
\multirow{4}{*}{NPI Licensing (own)}
 & 68M & 100 & 34.0 & 81.0 & $-47.0$ & 18.0 & 75.0 & $-57.0$ \\
 & 130M & 100 & 41.0 & 83.0 & $-42.0$ & 26.0 & 82.0 & $-56.0$ \\
 & 1.3B & 100 & 73.0 & 95.0 & $-22.0$ & 46.0 & 87.0 & $-41.0$ \\
 & 5.6B & 100 & 86.0 & 99.0 & $-13.0$ & 54.0 & 91.0 & $-37.0$ \\
\midrule
\multirow{4}{*}{Exist.-there Quant.}
 & 68M & 1000 & 91.7 & 91.7 & $+0.0$ & 68.5 & 68.5 & $+0.0$ \\
 & 130M & 1000 & 97.1 & 97.1 & $+0.0$ & 87.5 & 87.5 & $+0.0$ \\
 & 1.3B & 1000 & 98.1 & 98.1 & $+0.0$ & 93.4 & 93.4 & $+0.0$ \\
 & 5.6B & 1000 & 98.1 & 98.1 & $+0.0$ & 91.3 & 91.3 & $+0.0$ \\
\midrule
\multirow{4}{*}{Island Effects}
 & 68M & 1000 & 21.8 & 21.8 & $+0.0$ & 12.6 & 12.6 & $+0.0$ \\
 & 130M & 1000 & 22.0 & 22.0 & $+0.0$ & 19.5 & 19.5 & $+0.0$ \\
 & 1.3B & 1000 & 39.3 & 39.3 & $+0.0$ & 27.8 & 27.8 & $+0.0$ \\
 & 5.6B & 1000 & 58.7 & 58.7 & $+0.0$ & 44.5 & 44.5 & $+0.0$ \\
\midrule
\multirow{4}{*}{\textit{NPI Lic. (BLiMP)}}
 & \textit{68M} & \textit{7000} & \textit{39.0} & \textit{70.3} & \textit{$-31.3$} & \textit{51.5} & \textit{76.2} & \textit{$-24.7$} \\
 & \textit{130M} & \textit{7000} & \textit{47.3} & \textit{70.1} & \textit{$-22.8$} & \textit{63.0} & \textit{78.4} & \textit{$-15.4$} \\
 & \textit{1.3B} & \textit{7000} & \textit{64.1} & \textit{73.6} & \textit{$-9.5$} & \textit{66.5} & \textit{79.8} & \textit{$-13.3$} \\
 & \textit{5.6B} & \textit{7000} & \textit{72.2} & \textit{81.6} & \textit{$-9.4$} & \textit{71.9} & \textit{79.7} & \textit{$-7.8$} \\
\midrule
\midrule
\multicolumn{9}{l}{\textbf{\textit{BEAR facts}}} \\
\midrule
Wikipedia & 4.8B & 3221 & 20.5 & 7.9 & $+12.6$ & 17.3 & 7.1 & $+10.2$ \\
Common Corpus & 5.6B & 601 & 18.8 & 7.4 & $+11.4$ & 15.1 & 8.9 & $+6.2$ \\
\midrule
\midrule
\multicolumn{9}{l}{\textbf{\textit{GPT-2}} (pretrained, no No-$\mathcal{X}$ counterpart)} \\
 & & & \multicolumn{3}{c}{\textbf{GPT-2 small} (124M)} & \multicolumn{3}{c}{\textbf{GPT-2 medium} (355M)} \\
\cmidrule(lr){4-6}\cmidrule(lr){7-9}
Binding Reflexives & --- & 1000 & 67.3 & 67.3 & $+0.0$ & 68.6 & 68.6 & $+0.0$ \\
NPI Licensing (own) & --- & 100 & 91.0 & 97.0 & $-6.0$ & 92.0 & 97.0 & $-5.0$ \\
Exist.-there Quant. & --- & 1000 & 97.0 & 97.0 & $+0.0$ & 98.5 & 98.5 & $+0.0$ \\
Island Effects & --- & 1000 & 57.7 & 57.7 & $+0.0$ & 50.4 & 50.4 & $+0.0$ \\
\textit{NPI Lic. (BLiMP)} & --- & \textit{7000} & \textit{73.1} & \textit{67.7} & \textit{$+5.4$} & \textit{74.6} & \textit{70.9} & \textit{$+3.7$} \\
\midrule
BEAR (Wikipedia subset) & --- & 3221 & 11.6 & 4.8 & $+6.8$ & 18.3 & 5.1 & $+13.2$ \\
BEAR (Common Corpus subset) & --- & 601 & 28.1 & 9.0 & $+19.1$ & 39.6 & 10.0 & $+29.6$ \\
\bottomrule
\end{tabular*}
\caption{\textbf{Full-sentence log-likelihood (FS-LL) vs.\ length-normalised log-likelihood (LN-LL) accuracy (\%)}, for full models and their No-$\mathcal{X}$ (data-filtered) counterparts across training budgets. $N$ is the number of scored items per row: minimal pairs for the linguistic phenomena, corpus-supported facts for BEAR. $\Delta$ = FS-LL $-$ LN-LL (positive = FS-LL higher, negative = LN-LL higher). The final block reports the \texttt{gpt2} and \texttt{gpt2-medium} baselines, which have no data-filtered counterpart; there the two column groups are the two model sizes. For facts, we report both the $N=601$ corpus-supported facts in the Common Corpus set and the $N=3221$ in the Wikipedia set.}
\label{tab:fsll_vs_lnll}
\end{table*}

\autoref{tab:fsll_vs_lnll} reports model performance for both settings, linguistic phenomena and factual knowledge on both metric, summed and length-normalized sequence likelihood. \autoref{tab:prec10_fsll_vs_lnll} reports Precision@10 for the linguistic phenomena on queries the full model learned while the No-$\mathcal{X}$ model did not, i.e.\ those attributable to the target property. For \textit{existential there quantifiers} and \textit{wh-island effects} both verdicts select the identical query set. \textit{Binding reflexives} rise and \textit{NPIs} fall. Both follow the length bias discussed above: \emph{himself} is two tokens while the other reflexives are one, over-admitting those pairs under FS-LL; for NPIs the licensor lengthens the grammatical member, so the No-$\mathcal{X}$ model spuriously passes under LN-LL. The ranking of methods, however, is unchanged throughout.

\begin{table*}[!h]
\centering
\small
\setlength{\tabcolsep}{6pt}
\begin{tabular*}{\textwidth}{@{\extracolsep{\fill}} l rr rr rr rr @{}}
\toprule
& \multicolumn{2}{c}{$N$}
& \multicolumn{2}{c}{GradSim}
& \multicolumn{2}{c}{TrackStar}
& \multicolumn{2}{c}{K-FAC} \\
\cmidrule(lr){2-3}\cmidrule(lr){4-5}\cmidrule(lr){6-7}\cmidrule(lr){8-9}
Phenomenon & FS-LL & LN-LL & FS-LL & LN-LL & FS-LL & LN-LL & FS-LL & LN-LL \\
\midrule
\addlinespace[4pt]
Binding Reflexives  & 354 & 222 & 11.9 & 15.9 & 20.1 & 23.9 & 30.3 & 46.2 \\
\addlinespace[4pt]
NPI Licensing (own) &  33 &   8 & 34.5 & 16.2 & 45.2 & 22.5 & 50.9 & 27.5 \\
\addlinespace[4pt]
Exist.-there Quant. &  69 &  69 & 76.7 & 76.7 & 77.8 & 77.8 & 85.5 & 85.5 \\
\addlinespace[4pt]
Island Effects      & 191 & 191 & 37.1 & 37.1 & 15.4 & 15.4 & 50.4 & 50.4 \\
\addlinespace[4pt]
\bottomrule
\end{tabular*}
\caption{\textbf{Precision@10 of the influence contrast $\Delta\mathcal{I}$ under full-sentence (FS-LL) vs.\ length-normalised (LN-LL) likelihood scoring}, for the 5.6B Common Corpus model. $N$ is the number of queries, which are 'attributable' to $\mathcal{X}$, i.e.\ learned by the full model but not by the No-$\mathcal{X}$ model.}
\label{tab:prec10_fsll_vs_lnll}
\end{table*}

\paragraph{Learned- and not learned-definition.}

Our learned/not-learned definition (\autoref{eq:model_preference_check}) uses a
deliberately loose criterion: a model counts as having learned a query if it
merely \emph{prefers} the correct member, i.e.\ assigns it the higher summed
log-probability.

In the linguistic setting each query has a single
counterexample, so this is one preference check between the grammatical
sentence and its ungrammatical counterpart. In the factual setting each fact
comes with three templates and multiple answer options, so we require the
correct object to be the most likely candidate under all three templates; a
fact counts as not learned as soon as any single template ranks some
counterexample above the correct answer.

\begin{figure*}[!h]
    \centering
    \includegraphics[width=1\linewidth]{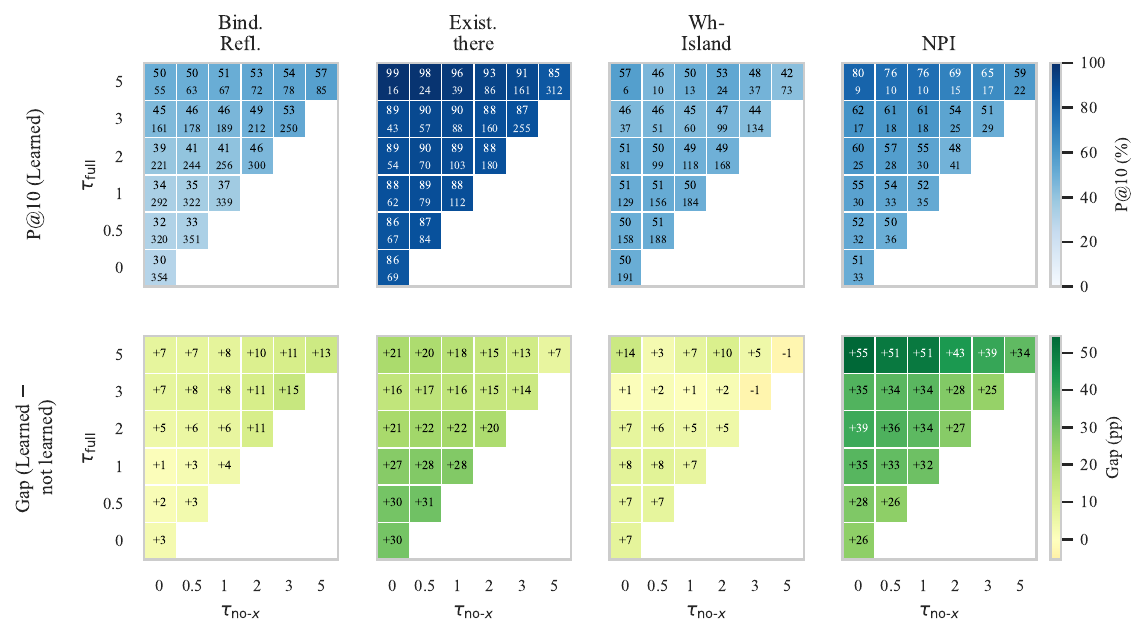}
\caption{\textbf{Linguistic phenomena: Precision@10 of the influence contrast
$\Delta\mathcal{I}$ on Learned vs.\ not learned queries for the 5.6B model on K-FAC approximation, across different verdicts of 'learning' vs 'not learned'.} We define thresholds $\tau \in \{0, 0.5, 1, 2, 3, 5\}$ for both the full model ($(\tau_{\text{full}}$) and the No-$\mathcal{X}$ $(\tau_{\text{no-}x})$ and ask how attribution results change for different margins for $(\tau_{\text{full}}$ and $\tau_{\text{no-}x})$. We require that $(\tau_{\text{full}} \geq (\tau_{\text{no-}x})$. $(\tau_{\text{full}} = \tau_{\text{no-}x} = 0$ represents the choice throughout the main parts of the paper. \emph{Top row:} Precision@10 on learned queries, which are 'attributable' to $\mathcal{X}$, and below the number of remaining items for each of the threshold. \emph{Bottom row:} the gap Precision@10(Learned) $-$ Precision@10(Not
learned), i.e, how much higher Precision@10-attribution to $\mathcal{X}$ is on learned queries compared to not learned queries.}
\label{fig:verdict_grid_ling}
\end{figure*}

\begin{figure*}[!h]
    \centering
    \includegraphics[width=0.65\linewidth]{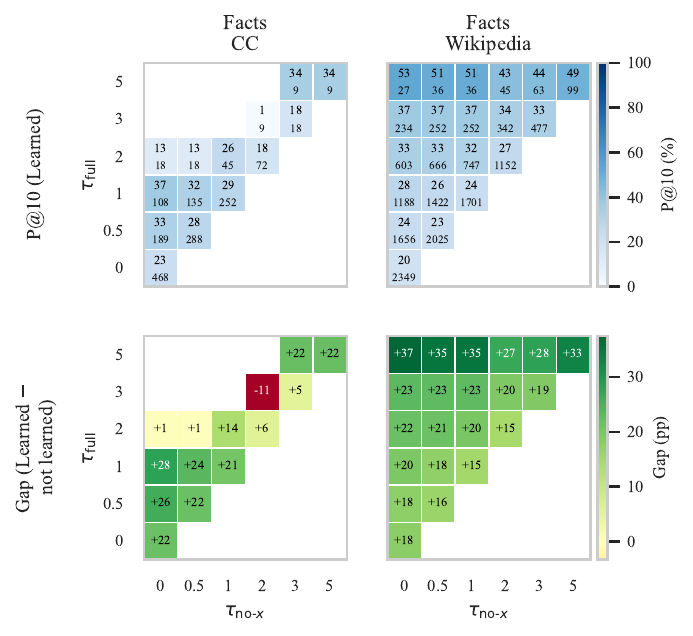}
\caption{\textbf{Facts: Precision@10 of the influence contrast $\Delta\mathcal{I}$
on learned vs.\ not learned queries, across the full verdict grid.} Same layout
as \autoref{fig:verdict_grid_ling}, for the two factual corpora (CC~5.6B and
Wikipedia).}
\label{fig:verdict_grid_facts}
\end{figure*}

This criterion is loose in two ways. At the item level the margin between
$s^+$ and $s^-$ may be small, so a `preference' need not reflect
`confidence'. And under filtering, a query with a large pre-filtering margin can
retain a clear (if reduced) preference for $s^+$ in the No-$\mathcal{X}$ model, yet still
be counted as \emph{learned} by both models.

To test whether this affects our conclusions, we re-run our analysis under
stricter verdicts that require a minimum margin $\tau$ between $s^+$ and $s^-$,
sweeping $\tau \in \{0, 0.5, 1, 2, 3, 5\}$ ($\tau{=}0$ recovers the main
results). 

To capture cases where filtering only partially reduces a large initial margin, we further allow separate thresholds for the two models, requiring $\tau_{\text{full}} \geq \tau_{\text{no-}x}$, so the full model must clear a higher bar than the No-$\mathcal{X}$ model. Tightening the margin necessarily discards queries that fall in the gap between the two thresholds.

\autoref{fig:verdict_grid_ling} and \autoref{fig:verdict_grid_facts} show how the attribution results change under different verdicts in the linguistic and factual setting. The top row of each plot reports Precision@$k$ of the influence contrast on queries the full model learned but the No-$\mathcal{X}$ model did not (those attributable to $\mathcal{X}$), together with the number of remaining queries per cell (below). The bottom row reports the gap in Precision@10 between these learned queries and not-learned ones. Its lower-left cell ($\tau_{\text{full}}{=}\tau_{\text{no-}x}{=}0$) reproduces the setup of the main paper.

The gap stays positive across nearly the entire grid, with two exceptions. For three of the four linguistic phenomena it is largest when the full-model threshold is high and the No-$\mathcal{X}$ threshold is low (upper-left cell). Similarly, for the facts and the Wikipedia model, the upper-left cell shows the largest gap. For the Common Corpus model no clear pattern emerges, mainly because few facts meet the threshold requirements (the low counts in the upper row of \autoref{fig:verdict_grid_facts}). We conclude that attribution is well-connected to acquisition. At the same time, for \textit{binding reflexives} and \textit{wh-island effects} the gaps are rather small, which we attribute to the errors discussed in \cref{sec:rq3}.

\section{Experimental Setup Details}
\label{sec:appendix_setup}

\subsection{Syntactically Annotated Corpus} \label{subsec:appendix_corpora}

\paragraph{Sources.} The use of fixed and re-shareable pre-training data is crucial to integrity and reproducibility, however, the licensing ambiguities of web-crawled data make this process exceedingly challenging. We build on the Common Corpus \citep{langlaisCommonCorpusLargest2026}---a multilingual collection of 2T tokens of uncopyrighted or permissable licensed text from public sources. It contains well-documented licensing and authorship metadata, which we maintain in our pre-processed version. 

\paragraph{Sampling \& Deduplication.} We applied a quality filter to the original Common Corpus data, removing any documents, which contain less than 85\% alphanumeric, or less than 85\% of the target language's characters (e.g., to avoid OCR artifacts). From the filtered data we sampled 10B tokens, sentence-segmented them, and parsed them (\emph{Parsing} below), yielding 2{,}565{,}931 documents and 340{,}886{,}709 sentences spanning 10{,}147{,}310{,}695 tokens of dependency-annotated text. We make the $\sim$ 10B dependency parsed corpus publicly available.\footnote{\url{https://huggingface.co/datasets/NLPnorth/under-the-influence}}

The authors of Common Corpus report a `negligible' amount of duplicates \cite{langlaisCommonCorpusLargest2026}. However, we observed that the resulting sample contains substantial near-verbatim repetition, e.g., boilerplate in patent claims and government filings. Since proper attribution to duplicate data points remains a significant problem \citep{patelWhenScoreBasedData2026}, we further deduplicated the sample. In particular, we hash each sentence of the sample by the MD5 digest of its space-joined \texttt{FORM} column (multiword-range and empty-node rows excluded) and only keep the first occurrence of each hash. 

We then split the deduplicated corpus into a 90\% training set and a 10\% test set, reserving the latter for future study of model behavior on unseen data. The models reported in this work are trained on the training set.

For the second corpus used in this work, the Wikipedia corpus, no such split is applied. In this case, we train the model on the full sentence-segmented corpus.

\begin{figure*}[!t]
    \centering
    \includegraphics[width=\linewidth]{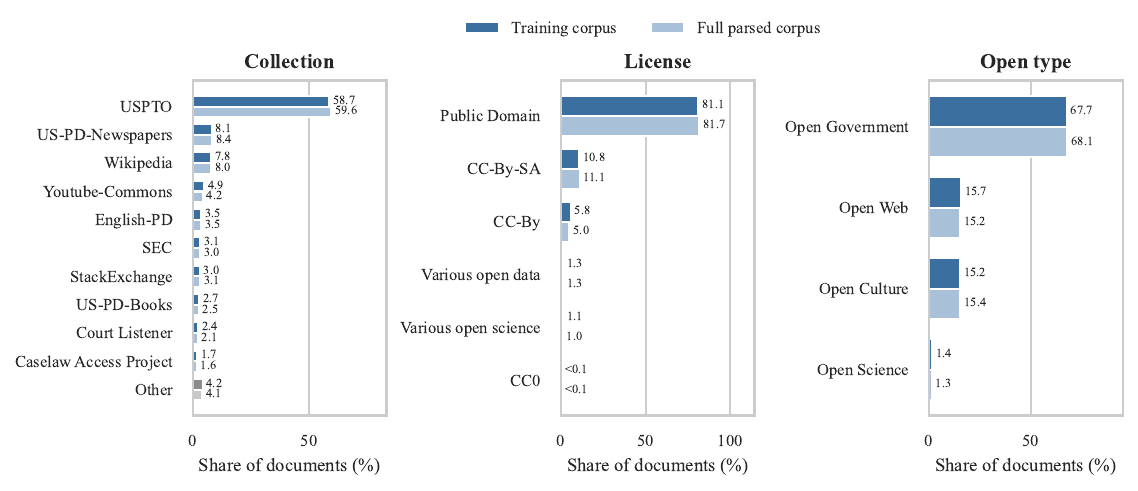}
    \caption{\textbf{Provenance of the full parsed corpus and of the corpus the models are trained on}, by source collection (top 10 of 47; the remaining 37 pooled as \emph{Other}), license, and open type.}
    \label{fig:corpus_composition}
\end{figure*}

\autoref{tab:appendix_corpus_stats} provides the amount of documents, sentences and words for each of stages of the pipeline, parsing, deduplication and train-test-splitting. \autoref{fig:corpus_composition} report the provenance metadata of that sample.

\begin{table}[!t]
\centering
\small
\setlength{\tabcolsep}{4pt}
\begin{tabularx}{\columnwidth}{@{}X r r@{}}
\toprule
 & \multicolumn{1}{c}{Count} & \multicolumn{1}{c}{\%} \\
\midrule
\multicolumn{3}{@{}l}{\textit{Documents}} \\
\quad Parsed corpus (released) & 2,565,931 &  \\
\quad Training corpus & 1,379,825 & 53.8 \\
\midrule
\multicolumn{3}{@{}l}{\textit{Sentences}} \\
\quad Parsed corpus (released) & 340,886,709 &  \\
\quad Training corpus & 162,691,844 & 47.7 \\
\quad\quad training split (90\%) & 146,422,036 & 43.0 \\
\quad\quad held-out split (10\%) & 16,269,808 & 4.8 \\
\midrule
\multicolumn{3}{@{}l}{\textit{Words (CoNLL-U tokens)}} \\
\quad Parsed corpus (released) & 10,147,310,695 &  \\
\quad Training corpus & 5,392,054,025 & 53.1 \\
\midrule
\multicolumn{3}{@{}l}{\textit{Tokens (Goldfish subwords)}} \\
\quad Training split & 5,660,798,493 &  \\
\bottomrule
\end{tabularx}
\caption{\textbf{Corpus sizes.} Percentages are shares of the released parsed
corpus. \emph{Training corpus} is what remains after deduplication; smaller budgets are random samples of it.}
\label{tab:appendix_corpus_stats}
\end{table}

\paragraph{Sentence length across the two corpora.} The Wikipedia corpus spans more sentences than the Common Corpus sample (195.6M vs.\ 146.4M) despite the smaller token budget (4.8B vs.\ 5.6B), because Common Corpus sentences are considerably longer: 38.7 tokens on average against Wikipedia's 24.3. This follows from the composition reported above. Patent claims are the largest component of the sample, which are single-sentence constructions of exceptional length. 

\paragraph{Parsing} For our linguistic experiments, we used MaChAmp v0.4.2~\cite{van-der-goot-etal-2021-massive} with default hyperparameters, train a single multi-task model that performs word segmentation, UPOS tagging, XPOS tagging, lemmatization, morphological labeling, and dependency parsing. We trained parsers based on four different language models: deberta-v3-large~\cite{he2023debertav}, luke-large~\cite{yamada-etal-2020-luke}, roberta-large~\cite{liuRoBERTaRobustlyOptimized2019}, ModernBERT-large~\cite{warner-etal-2025-smarter}. We manually inspected 20 sentences on which the parsers disagreed, i.e., where at least two models produced different dependency structures, and found the deberta-based parser to be the most accurate.

\subsection{Models and Hyperparameters}
\label{subsec:appendix_models}

\begin{table}[]
\centering
\small
\setlength{\tabcolsep}{4pt}
\begin{tabularx}{\columnwidth}{@{}X r r@{}}
\toprule
 & \textbf{Small (ours)} & \textbf{Medium (ours)} \\
\midrule
Layers $L$                      & 12         & 24 \\
Attention heads $H$             & 12         & 16 \\
Hidden size $d_{\text{model}}$  & 768        & 1{,}024 \\
FFN size $d_{\text{ff}}$        & 3{,}072    & 4{,}096 \\
Context length $n_{\text{ctx}}$ & 512        & 1{,}024 \\
Vocabulary                      & 51{,}200   & 51{,}200 \\
\midrule
Parameters           & 124.8M     & 355.8M \\
\midrule
Activation                      & GELU       & GELU \\
Dropout (attn/emb/resid)        & 0.1        & 0.1 \\
LayerNorm $\varepsilon$         & $10^{-5}$  & $10^{-5}$ \\
Init.\ std.                     & 0.02       & 0.02 \\
Tied input/output embeddings    & yes        & yes \\
\midrule
Used for budgets                & $\leq$1.3B & $>$1.3B \\
\bottomrule
\end{tabularx}
\caption{\textbf{Model architectures.} Both are standard GPT-2 configurations
\citep{radfordLanguageModelsAre} with the vocabulary set by the Goldfish
tokenizer \citep{changGoldfishMonolingualLanguage2024}, a SentencePiece unigram
model with 51{,}200 types trained on 130MB of English. Sequences are tokenized
to at most 512 tokens in both cases.}
\label{tab:appendix_architecture}
\end{table}

\paragraph{Architecture.}
\autoref{tab:appendix_architecture} lists the two architectures we use. For
token budgets up to 1.3B we use the GPT-2 small configuration, and GPT-2 medium
\citep{radfordLanguageModelsAre} for larger budgets. All models are trained from
scratch. We train our own SentencePiece unigram tokenizer on a 130MB sample of our Common Corpus subset, following the Goldfish recipe \citep{changGoldfishMonolingualLanguage2024}.
We do not retrain it for different token budgets, nor for the Wikipedia-based model.

\paragraph{Optimization.}
All runs share one recipe, taken from Goldfish
\citep{changGoldfishMonolingualLanguage2024}: AdamW
($\beta_1=0.9$, $\beta_2=0.999$) with learning rate $1\times10^{-4}$,
$\varepsilon = 10^{-6}$, weight decay $0.01$, gradient clipping at $1.0$, linear
decay to zero after a linear warmup over the first $10\%$ of steps, and
\texttt{bf16} mixed precision. Every model is trained for exactly one epoch. The
effective batch size is 64 sequences for the small models and 32 for the medium
models. Random seeds are fixed per budget and shared between a full model and its No-$\mathcal{X}$ counterpart.

\paragraph{Data handling.}
The corpus is sentence-segmented (app.~\ref{subsec:appendix_corpora}), tokenized
into sequences of at most 512 tokens, and \emph{shuffled at the sequence level}, and
Shuffling is done independently per model, so a full model and its
No-$\mathcal{X}$ counterpart do not share an example ordering. Importantly, every
(budget, phenomenon) pair is trained \emph{from scratch} rather than branched
from a shared checkpoint to account for the sensitivity of IFs to training stochasticity \citep{mlodozeniecDistributionalTrainingData2025}.

\paragraph{Baselines.}
The pretrained \texttt{gpt2} (124M) and \texttt{gpt2-medium} (355M)
checkpoints\footnote{\url{https://huggingface.co/openai-community/gpt2}} are used as released, with their own BPE tokenizer and 1{,}024-token context. They are evaluated but never retrained, and have no No-$\mathcal{X}$ counterpart.

\begin{table}[]
\centering
\small
\setlength{\tabcolsep}{4pt}
\begin{tabularx}{\columnwidth}{
  @{}>{\raggedright\arraybackslash}X r@{}
}
\toprule
\multicolumn{2}{@{}l}{\textbf{\textit{Shared (except BM25)}}} \\
Max.\ sequence length             & 512 \\
Random projection (per module)    & $64 \times 64$ \\
Projection type                   & Rademacher \\
Two-sided (block-diagonal)        & yes \\
Unit normalization (cosine)       & yes \\
Modules scored                    & all linear \\
\quad excluded                    & \texttt{lm\_head}, embeddings \\
\midrule
\multicolumn{2}{@{}l}{\textbf{\textit{GradSim}}} \\
Optimizer-state normalization     & off \\
\midrule
\multicolumn{2}{@{}l}{\textbf{\textit{TrackStar}}} \\
Downweighted components           & 1{,}000 \\
Optimizer-state (Adafactor) term  & off \\
\midrule
\multicolumn{2}{@{}l}{\textbf{\textit{K-FAC}}} \\
Damping $\lambda$                 & 0.1 \\
Application scheme                & fused (iii) \\
\midrule
\midrule
\multicolumn{2}{@{}l}{\textbf{\textit{BM25}}} \\
Stopwords                         & retained \\
\bottomrule
\end{tabularx}
\caption{\textbf{Attribution hyperparameters}, shared across both settings and
all token budgets. `Fused' refers to scheme (iii) of
Appendix~\ref{subsec:appendix_curvature}, in which whitening is folded into the
projection matrices.}
\label{tab:appendix_attribution_hparams}
\end{table}

\paragraph{Attribution hyperparameters.}
\autoref{tab:appendix_attribution_hparams} lists the attribution settings, held fixed across phenomena, corpora, and token budgets. Gradients are compressed with a two-sided Rademacher projection to $64 \times 64$ per weight module and unit-normalized (app.~\ref{subsec:appendix_cosine_projections}), so all three influence methods return cosine-similarity scores on a common scale; BM25 scores are $L_2$-normalized per query for the same reason. 

The language-model head and the token embeddings are excluded from all approximation methods. K-FAC uses the fused scheme (iii) detailed in app.~\ref{subsec:appendix_curvature} with damping $\lambda = 0.1$. Note, that for K-FAC damping is applied in \texttt{auto\_mean} mode, following \citet{quirkeBergsonOpenSource2026}: $\lambda_{\text{real}} = \lambda \cdot \sigma(\hat{H})$, the mean eigenvalue of the approximation scaled by the multiplier $\lambda$.

\paragraph{Compute.}
All experiments ran on a single GPU per job, either an NVIDIA H100 NVL or an L40S. Total GPU-hours, energy, and emissions are reported in
\cref{sec:ethics}.

\subsection{Construction of Filters}
\label{appendix:filters}

\subsubsection{Linguistic Setting}
\label{subsec:appendix_linguistic}

We select phenomena while aiming to balance several desiderata: (1) easy to isolate, unlikely to be learned by similar constructions, (2) a small enough filtered subset to keep the training data largely undisturbed, (3) spanning different levels of syntactic structure. This section describes each phenomenon and how its filter is applied.

\begin{itemize}[leftmargin=1em, itemsep=2pt, topsep=2pt]

  \item \textbf{Binding reflexives} hinge on
  single closed-class lexical items. Here we experimented with lemma based filters targeting the parsed data's lemma field, but manual checks on filtered results showed that the recall was low. As such, we filtered them based on surface form matches. Evaluation is conducted over a single BLiMP suite, \texttt{principle\_A\_c\_command}, which solely targets reflexive binding.

 \item \textbf{NPIs.} Similar to \textit{binding}, we first experimented with syntactic filters on dependency parses, targeting only the licensed NPIs and letting unlicensed and polysemous forms pass through. However, manual checks showed that parser accuracy deteriorated with longer dependencies, such as NPIs licensed through matrix clauses. Again erring on the side of caution, we filtered them based on surface form matches. For evaluation, we use a subset of BLiMPPI (app.~\ref{subsec:appendix_blimp_npi}).

\item \textbf{\emph{Wh}-island effects} are tied to a sentence type that is very rare in written corpora, where wh-words are followed by a nominal, while preserving the question form. We use syntactic filters to isolate the sentences exhibiting this structure. The BLiMP suite \texttt{left\_branch\_island\_simple\_question} is used for evaluation.

  \item \textbf{Existential there quantifiers} combine a frequent construction, \textit{existential there}, with a closed set of quantifier function words with narrowly defined semantics, and thus test the regime where $\mathcal{X}$ is common but still lexically bounded. Again we use syntactic filters, targeting existentials governing a quantifier. For evaluation, \texttt{existential\_there\_quantifiers\_1} and \texttt{\_2} are used.

\end{itemize}

\paragraph{Filter Precision and Recall}
We sampled 100 candidate sentences per phenomenon with a lexical trigger, annotated them by hand, and ran the filter over the same parses. ETQ: precision 0.93, recall 0.78. Wh-islands: precision 0.62, recall 0.62. For binding and NPIs the filters are lexical rather than syntactic, so we did not evaluate them separately.

Most misses stem from broken parses rather than from the filter patterns. The missed sentences typically lack whitespace or are otherwise corrupted, so the correct parses never exist. Imperfect recall means some deformed forms of $\mathcal{X}$ still survive in $\mathcal{D}_{base}$.\footnote{Annotations are released with our code.}

\subsubsection{Why Not BLiMP's NPI Set?}
\label{subsec:appendix_blimp_npi}
Two types of contrast in BLiMP's NPI suite proved incompatible with our framework. In \texttt{npi\_present\_1} and \texttt{npi\_present\_2}, a grammatical positive sentence is contrasted with an ungrammatical counterpart in which an adverb is replaced by an NPI: \expected{\emph{Many skateboards also roll.}} vs.\ \unexpected{\emph{*Many skateboards ever roll.}} Since the ablated models are trained without NPIs, the ungrammatical member,
containing a word form the model has never seen, becomes even less likely. As shown in \autoref{tab:filtering_linguistic_full}, No-$\mathcal{X}$ models consequently score higher than the full models.

The second contrast we observed issues with is
\texttt{only\_npi\_licensor\_present}, which keeps the NPI constant and varies the presence of the licensor (\expected{\emph{Only Peter ever comes here.}}
vs.\ \unexpected{\emph{*Even Peter ever comes here.}}). Here the licensor \textit{only} is always contrasted with the non-licensing adverb \textit{even}. As discussed in \cref{appendix:metric_verdict}, the two are
tokenized differently, which confounds the contrast. The remaining suites, which keep the NPI constant and vary either the scope or the presence of the licensor, are technically viable. However, BLiMP's NPI suites cover a single NPI, \textit{ever}, so they test whether models learn the distribution of \textit{ever}, rather than NPIs as a phenomenon.

We therefore use BLiMPPI \citep{güven2026blimppibenchmarklinguisticminimal}, specifically the portion derived from Universal Dependencies \citep{nivre-etal-2020-universal}. This provides naturally occurring instances of different NPIs: \textit{anything}, \textit{anywhere}, \textit{ever}, \textit{anymore},
\textit{yet}, \textit{at all}, \textit{either}, and \textit{whatsoever}. The licensing environments are also more varied: sentential negation, \textit{n}-series negation
(e.g.\ \textit{nobody}), downward-entailing operators (e.g.\
\textit{hardly}), and questions. We use 100 probes from BLiMPPI, released alongside our codebase.

\subsubsection{Factual Setting}
\label{subsec:appendix_factual}

\begin{table}[!t]
  \centering
  \scriptsize
  \begin{tabularx}{\columnwidth}{
    @{}>{\raggedright\arraybackslash}X r r r r@{}
  }
    \toprule
    \multirow{2}{*}{Corpus}
      & \multirow{2}{*}{Filtered out}
      & \multicolumn{2}{c}{Sentences}
      & \multirow{2}{*}{$\Delta$} \\
    \cmidrule(lr){3-4}
      & & Full & No-$\mathcal{X}$ & \\
    \midrule
    CC 5.6B
      & 891{,}011
      & 146{,}422{,}036
      & 145{,}531{,}025
      & $-0.61\%$ \\
    Wikipedia
      & 2{,}349{,}171
      & 195{,}620{,}725
      & 193{,}271{,}554
      & $-1.20\%$ \\
    \bottomrule
  \end{tabularx}
  \caption{\textbf{Training-set size in the factual setting.}
  Number of sentences filtered out and number of sentences remaining in the
  training corpus for each full model and its No-$\mathcal{X}$ counterpart.
  $\Delta$ is computed within a corpus, so the two rows are directly comparable
  as filter costs but not as corpus sizes: the Common Corpus figure is the 90\%
  training split (\autoref{subsec:appendix_corpora}) and its sentences are
  roughly $1.6\times$ longer than Wikipedia's, so the same sentence percentage
  corresponds to a larger share of tokens there.}
  \label{tab:facts_filter_cost}
\end{table}

This section details the factual filtering procedure. \autoref{tab:facts_filter_cost} reports the amount of data removed.

\paragraph{Restricting to corpus-supported relations.}
Some BEAR relations \citep{wilandBEARUnifiedFramework2024} are answered
through prior bias rather than knowledge, e.g., when one candidate object is
frequent while the others never appear. For $5{,}938$ of the $7{,}731$ relations ($76.8\%$) in the 5.6B Common Corpus subset, and
for $1{,}875$ ($24.3\%$) in the Wikipedia, the corresponding entities don't co-occur in a single sentence in the corpus. We therefore restrict the analysis to \emph{corpus-supported} relations, whose two entities co-occur in at
least $10$ sentences of the training corpus. This leaves $601$ relations at
5.6B and $3{,}221$ for Wikipedia (\autoref{tab:filtering_facts_full}).

\paragraph{Which relations are filtered?}
We filter only relations that are corpus-supported and answered correctly by
the full model $f_\theta$ (correct object ranked first on all three
templates). Filtering the not-learned relations would shrink the corpus
without changing anything the No-$\mathcal{X}$ model can be asked about.

\paragraph{What is removed?}
For each such relation we remove every sentence where the subject and object
co-occur. Following \citet{nadkarniRewritingHistoryRecipe2026}, who find that
this alone suppresses fact learning only partially, we also remove
subject-only and object-only sentences. This step is applied per entity, and
only to entities occurring in fewer than $30{,}000$ corpus sentences. The cap
prevents deleting a large, topically coherent slice of the corpus for a
high-frequency entity such as \emph{Europe}, which would confound the
$f_\theta$ vs.\ $f_{\theta^-}$ comparison with a general distribution shift.

\paragraph{Surface forms and matching.}
Entities take many surface forms, so for each we collect the union of (i) its
Wikidata label, aliases, and demonyms via the Wikidata API
\citep{vrandecicWikidataFreeCollaborative2014}; (ii) aliases shipped with
BEAR; and (iii) a manually curated table of country names, demonyms, and
capital names. Matching is at whole-word boundaries, case-insensitive with a
Unicode (NFKD) fallback, except for short all-caps acronyms (\emph{US},
\emph{UK}, \emph{DC}), which are matched case-sensitively.

\paragraph{Inseparable entity pairs.}
Some pairs cannot be told apart by any surface-form filter. We drop a relation
if its both entities share a surface form after surface matching
(\emph{Italy--Italian}), or if one canonical label contains the other as a
whole word (\emph{Mexico City--capital--Mexico}); the containment test
requires at least four characters. This removes $13$ relations in each of the
Common Corpus 5.6B and Wikipedia settings.

\begin{table}[!t]
\centering
\small
\begin{tabular}{@{}llrrrr@{}}
\toprule
Phenomenon & Budget & $n$ & Exact & Parad. & None \\
\midrule
\multirow{4}{*}{\makecell[l]{Exist.-there\\Quantifiers}}
 & 68M   & 6,666 & 91.7 &  8.1 & 0.2 \\
 & 130M  & 7,560 & 94.7 &  5.3 & 0.0 \\
 & 1.3B  & 7,575 & 95.3 &  4.6 & 0.1 \\
 & 5.6B  & 7,722 & 88.5 & 11.3 & 0.2 \\
\midrule
\multirow{4}{*}{\makecell[l]{Binding\\Reflexives}}
 & 68M   & 7,279 & 99.0 &  0.8 & 0.2 \\
 & 130M  & 7,361 & 99.2 &  0.7 & 0.1 \\
 & 1.3B  & 5,019 & 98.9 &  1.1 & 0.0 \\
 & 5.6B  & 4,448 & 96.3 &  3.5 & 0.2 \\
\midrule
\multirow{4}{*}{\makecell[l]{NPIs}}
 & 68M   &   233 & 69.5 & 24.0 & 0.4 \\
 & 130M  &   216 & 71.8 & 23.6 & 0.0 \\
 & 1.3B  &   326 & 76.1 & 14.7 & 0.3 \\
 & 5.6B  &   310 & 79.0 & 17.4 & 0.6 \\
\midrule
\multirow{4}{*}{\makecell[l]{Wh-island\\effects}}
 & 68M   & 3,187 & 92.1 &  6.1 &  1.8 \\
 & 130M  & 4,086 & 85.0 &  2.1 & 12.9 \\
 & 1.3B  & 4,302 & 98.6 &  1.4 &  0.0 \\
 & 5.6B  & 4,197 & 98.7 &  1.2 &  0.1 \\
\bottomrule
\end{tabular}
\caption{\textbf{Diversity of target hits} (K-FAC, retrievals in
$\mathcal{D}_\mathcal{X}$). \emph{Exact}: the retrieval reuses the probe's own
item; \emph{Parad.}: a different member of the same paradigm; \emph{None}: no
item found. All values in \%; rows do not sum to 100 where the probe's own item
was not matched by the extractor (NPIs, 3--9\%). $n$ is the number of retrieved
training points scored.}
\label{tab:diversity}
\end{table}

\onecolumn
\section{Supplementary Experimental Results}

\label{sec:appendix_additional_results}

\begin{table*}[b]
  \centering
  \scriptsize
  \setlength{\tabcolsep}{2.5pt}
  \begin{tabular*}{\textwidth}{@{\extracolsep{\fill}} ll rcc rcccc rrrr @{}}
    \toprule
    \multirow{2}{*}{Corpus} & \multirow{2}{*}{Tokens} & \multicolumn{3}{c}{Full BEAR} & \multicolumn{9}{c}{Corpus-supported} \\
    \cmidrule(lr){3-5}\cmidrule(l){6-14}
    & & $N$ & Acc & Rand & $N$ & Full & No-$\mathcal{X}$ & $\Delta$ & Rand & \makecell{both\\models wrong} & \makecell{Full correct,\\No-$\mathcal{X}$ wrong} & \makecell{both\\models correct} & \makecell{Full wrong,\\No-$\mathcal{X}$ correct} \\
    \midrule
    \multirow{2}{*}{\textit{Common Corpus}} & \textit{1.3B} & \textit{7731} & \textit{3.6\%} & \textit{0.04\%} & \textit{309} & \textit{23.6\%} & \textit{14.6\%} & \textcolor{green!50!black}{\textit{$-9.1\%$}} & \textit{0.15\%} & \textit{224} & \textit{40} & \textit{33} & \textit{12} \\
     & 5.6B & 7731 & 3.6\% & 0.04\% & 601 & 18.8\% & 15.1\% & \textcolor{green!50!black}{$-3.7\%$} & 0.10\% & 458 & 52 & 61 & 30 \\
    \midrule
    Wikipedia & 4.8B & 7731 & 13.4\% & 0.04\% & 3221 & 20.5\% & 17.3\% & \textcolor{green!50!black}{$-3.2\%$} & 0.04\% & 2404 & 261 & 398 & 158 \\
    \bottomrule
  \end{tabular*}
\caption{\textbf{Effect of data filtering on factual knowledge (BEAR).} A fact is a relation between two entities, e.g.\ \emph{The capital of X is Y.} The model answers by ranking the candidate options for one entity (Y) by per-template log-likelihood. We count a fact as \emph{learned} if the correct object ranks first under \emph{all three} of the relation's templates. Each row reports one training corpus and token budget under two evaluation sets. \emph{Full BEAR} is the entire benchmark ($N{=}7{,}731$ facts). \emph{Corpus-supported} restricts to the \emph{learnable} subset: facts whose two entities co-occur $\geq 10$ times in the training corpus and do not share a Wikidata alias or substring; $N$ is that subset's size. For the corpus-supported set, \emph{Full} and \emph{No-$\mathcal{X}$} are each model's fraction of learned facts, and $\Delta=\text{No-$\mathcal{X}$}-\text{Full}$ (\expected{green (-)}: $\Delta<0$, Full exceeds No-$\mathcal{X}$; \unexpected{red (+)}: $\Delta>0$, No-$\mathcal{X}$ exceeds full). \emph{Rand} is the chance level, averaging $(1/A)^3$ over the facts, where $A$ is the number of candidate objects in a fact's relation and the cube reflects independent uniform guesses being correct on all three templates. The final four columns partition the corpus-supported facts by the joint outcome of the full model and the No-$\mathcal{X}$ model (trained with the BEAR-supporting sentences removed).}%
    \label{tab:filtering_facts_full}
\end{table*}

\begin{table*}[!h]
  \centering
  \scriptsize
  \setlength{\tabcolsep}{4pt}
  \resizebox{\textwidth}{!}{%
  \begin{tabular*}{\textwidth}{@{\extracolsep{\fill}} ll r ccc r rrrr @{}}
    \toprule
    \multirow{2}{*}{Phenomenon} & \multirow{2}{*}{Tokens} & \multirow{2}{*}{$N$} & \multirow{2}{*}{Rand.} & \multirow{2}{*}{Full} & \multirow{2}{*}{No-$\mathcal{X}$} & \multirow{2}{*}{$\Delta$} & \multicolumn{4}{c}{Item counts} \\
    \cmidrule(l){8-11}
    & & & & & & & \multicolumn{1}{c}{both models}  & \multicolumn{1}{c}{Full correct,}  & \multicolumn{1}{c}{both models}  & \multicolumn{1}{c}{Full wrong,}  \\
    & & & & & & &  \multicolumn{1}{c}{wrong} & \multicolumn{1}{c}{No-$\mathcal{X}$ wrong} &  \multicolumn{1}{c}{correct} & \multicolumn{1}{c}{No-$\mathcal{X}$ correct} \\
    \midrule
    Binding Reflexives & 68M & 1000 & 50.0\% & 58.1\% & 61.7\% & \textcolor{red!70!black}{$+3.6\%$} & 107 & 276 & 305 & 312 \\
     & 130M & 1000 & 50.0\% & 59.6\% & 56.9\% & \textcolor{green!50!black}{$-2.7\%$} & 179 & 252 & 344 & 225 \\
     & 1.3B & 1000 & 50.0\% & 74.7\% & 30.5\% & \textcolor{green!50!black}{\textbf{$-44.2\%$}} & 196 & 499 & 248 & 57 \\
     & 5.6B & 1000 & 50.0\% & 75.0\% & 49.9\% & \textcolor{green!50!black}{$-25.1\%$} & 147 & 354 & 396 & 103 \\
    \addlinespace[2pt]
    NPI Lic. (own) & 68M & 100 & 50.0\% & 34.0\% & 18.0\% & \textcolor{green!50!black}{$-16.0\%$} & 65 & 17 & 17 & 1 \\
     & 130M & 100 & 50.0\% & 41.0\% & 26.0\% & \textcolor{green!50!black}{$-15.0\%$} & 57 & 17 & 24 & 2 \\
     & 1.3B & 100 & 50.0\% & 73.0\% & 46.0\% & \textcolor{green!50!black}{$-27.0\%$} & 26 & 28 & 45 & 1 \\
     & 5.6B & 100 & 50.0\% & 86.0\% & 54.0\% & \textcolor{green!50!black}{\textbf{$-32.0\%$}} & 13 & 33 & 53 & 1 \\
    \addlinespace[2pt]
    Exist.-there Quantifiers & 68M & 1000 & 50.0\% & 91.7\% & 68.5\% & \textcolor{green!50!black}{\textbf{$-23.2\%$}} & 83 & 232 & 685 & 0 \\
     & 130M & 1000 & 50.0\% & 97.1\% & 87.5\% & \textcolor{green!50!black}{$-9.6\%$} & 29 & 96 & 875 & 0 \\
     & 1.3B & 1000 & 50.0\% & 98.1\% & 93.4\% & \textcolor{green!50!black}{$-4.7\%$} & 19 & 47 & 934 & 0 \\
     & 5.6B & 1000 & 50.0\% & 98.1\% & 91.3\% & \textcolor{green!50!black}{$-6.8\%$} & 18 & 69 & 912 & 1 \\
    \addlinespace[2pt]
    Wh-island effects & 68M & 1000 & 50.0\% & 21.8\% & 12.6\% & \textcolor{green!50!black}{$-9.2\%$} & 739 & 135 & 83 & 43 \\
     & 130M & 1000 & 50.0\% & 22.0\% & 19.5\% & \textcolor{green!50!black}{$-2.5\%$} & 737 & 68 & 152 & 43 \\
     & 1.3B & 1000 & 50.0\% & 39.3\% & 27.8\% & \textcolor{green!50!black}{$-11.5\%$} & 579 & 143 & 250 & 28 \\
     & 5.6B & 1000 & 50.0\% & 58.7\% & 44.5\% & \textcolor{green!50!black}{\textbf{$-14.2\%$}} & 364 & 191 & 396 & 49 \\
    \midrule
    \textit{NPI Lic. (BLiMP)} & \textit{68M} & \textit{7000} & \textit{50.0\%} & \textit{39.0\%} & \textit{51.5\%} & \textcolor{red!70!black}{\textit{$+12.4\%$}} & \textit{2072} & \textit{1326} & \textit{1406} & \textit{2196} \\
     & \textit{130M} & \textit{7000} & \textit{50.0\%} & \textit{47.3\%} & \textit{63.0\%} & \textcolor{red!70!black}{\textbf{\textit{$+15.7\%$}}} & \textit{1367} & \textit{1224} & \textit{2084} & \textit{2325} \\
     & \textit{1.3B} & \textit{7000} & \textit{50.0\%} & \textit{64.1\%} & \textit{66.5\%} & \textcolor{red!70!black}{\textit{$+2.4\%$}} & \textit{841} & \textit{1502} & \textit{2985} & \textit{1672} \\
     & \textit{5.6B} & \textit{7000} & \textit{50.0\%} & \textit{72.2\%} & \textit{71.9\%} & \textcolor{green!50!black}{\textit{$-0.3\%$}} & \textit{547} & \textit{1419} & \textit{3633} & \textit{1401} \\
    \bottomrule
  \end{tabular*}}
  \caption{\textbf{Effect of data filtering on linguistic phenomena.} \emph{Full} is the accuracy of the model trained on the complete corpus; \emph{No-$\mathcal{X}$} is the accuracy of the model trained on the same corpus with the sentences supporting that phenomenon removed; both are full-sentence log-likelihood accuracy. \emph{Rand.}\ is the chance level of a two-alternative forced choice ($50\%$). $\Delta=\text{No-$\mathcal{X}$}-\text{Full}$ (\expected{green (-)}: $\Delta<0$, Full exceeds No-$\mathcal{X}$; \unexpected{red (+)}: $\Delta>0$, No-$\mathcal{X}$ exceeds full. $N$ is the number of minimal pairs. We additionally break down results by the joint correctness of the two models. Counter-intuitively, for binding reflexives and wh-island effects, the full model sometimes fails where the filtered model succeeds. For wh-islands this could be explained by the overall low performance. For binding reflexives it is unclear why this happens.}
\label{tab:filtering_linguistic_full}
\end{table*}

\begin{figure*}[!h]
    \centering
    \includegraphics[width=1\linewidth]{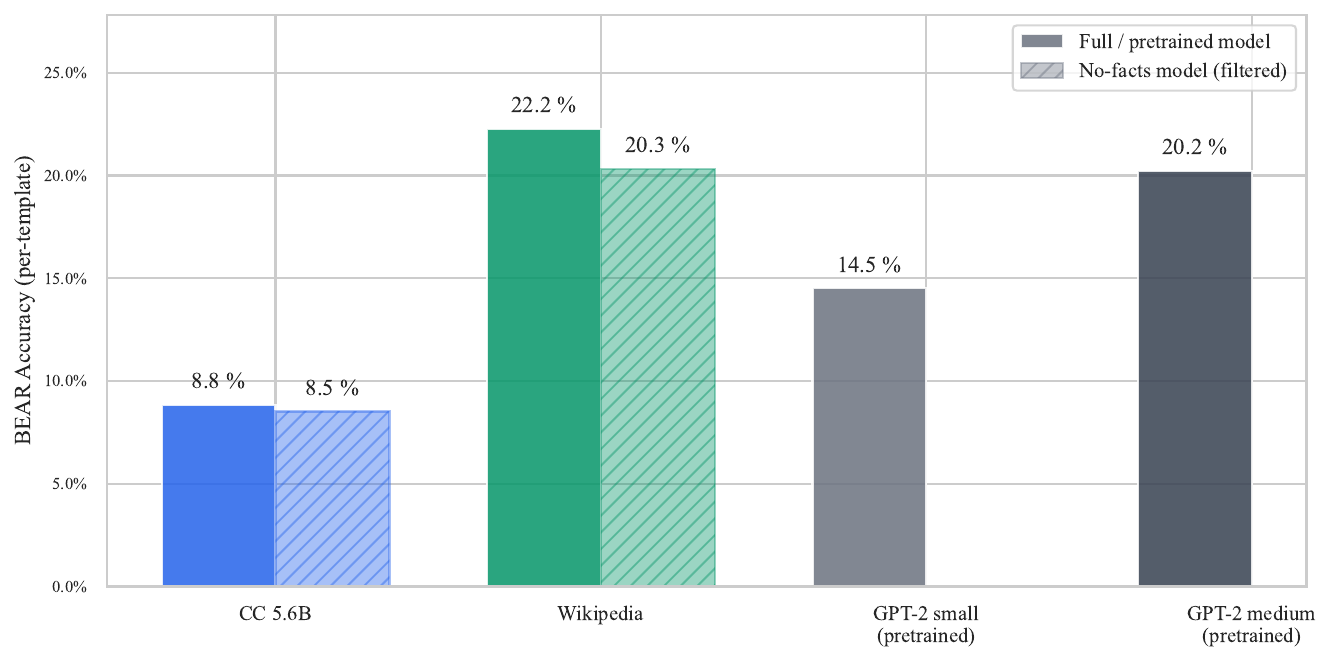}
\caption{Standard BEAR evaluation under the per-template perspective. BEAR \citep{wilandBEARUnifiedFramework2024} is built around templates, aggregating over each fact (a relation between two entities) and its templates. We instead adopt a relation-level perspective, reporting results at the level of facts. This limits comparability with standard benchmark results, so for reference this figure reports the per-template perspective, as used by \citet{warstadtBLiMPBenchmarkLinguistic2023} and the \textsc{LM-Pub-Quiz} leaderboard (\url{https://lm-pub-quiz.github.io/}).}
\label{fig:standard-bear-results}
\end{figure*}

\begin{figure*}[!h]
    \centering
    \includegraphics[width=1\linewidth]{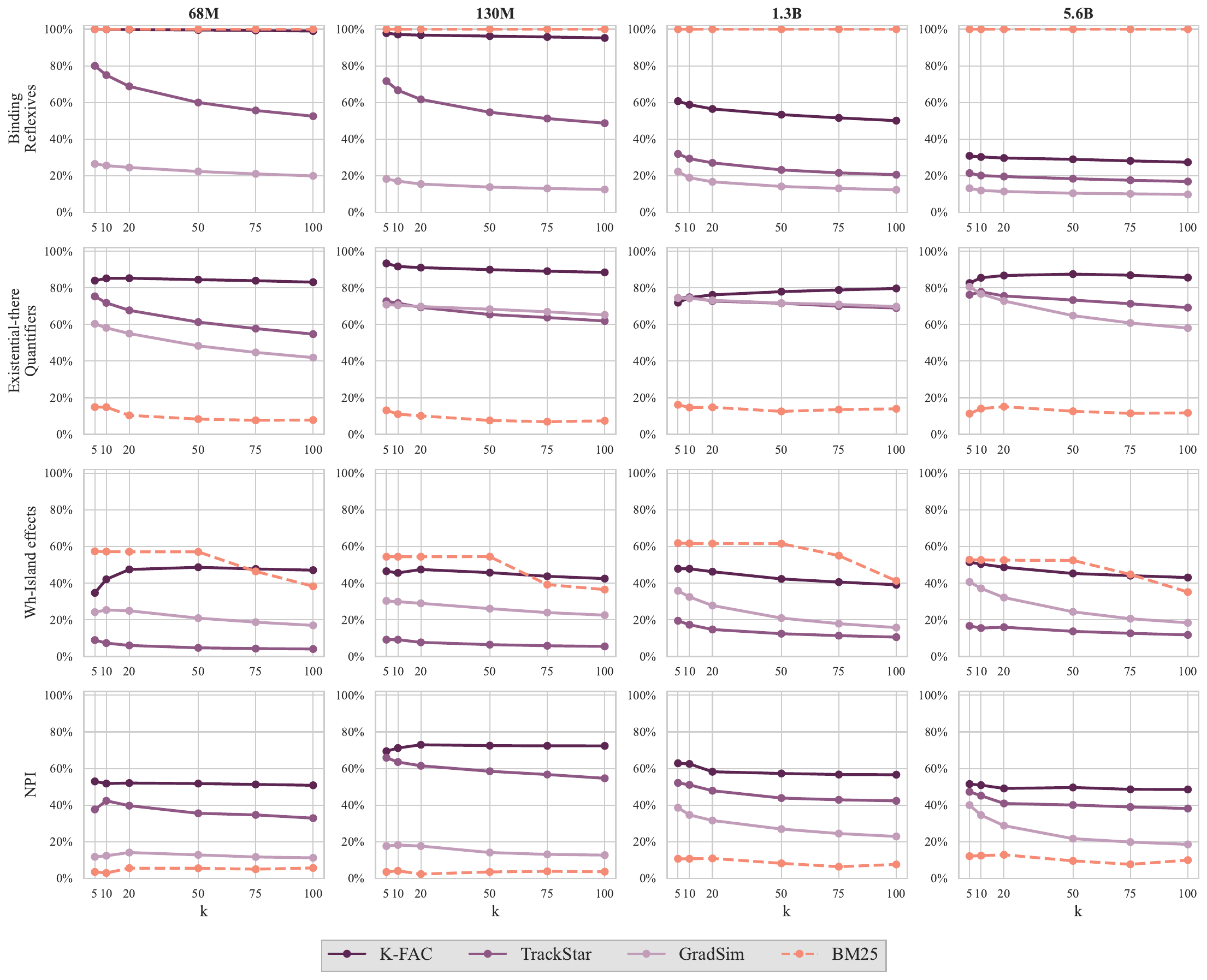}
\caption{\textbf{Precision@k of the influence contrast $\Delta \mathcal{I}$ across token budgets (68M--5.6B)}, for each linguistic phenomenon (rows) and attribution method.}
\label{fig:prec_k_full_linguistics}
\end{figure*}

\begin{figure*}[!h]
    \centering
    \includegraphics[width=0.7\linewidth]{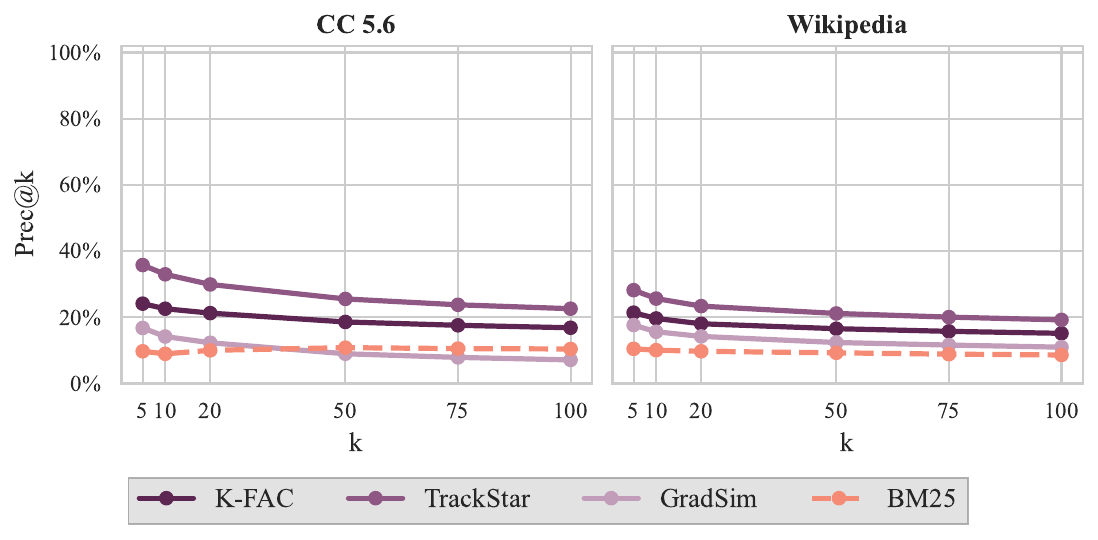}
\caption{\textbf{Precision@k of the influence contrast $\Delta \mathcal{I}$ for facts}, across attribution methods and both training corpora (Common Corpus 5.6B and Wikipedia).}
\label{fig:prec_k_full_facts}
\end{figure*}

\begin{figure*}[!h]
    \centering
    \includegraphics[width=0.7\linewidth]{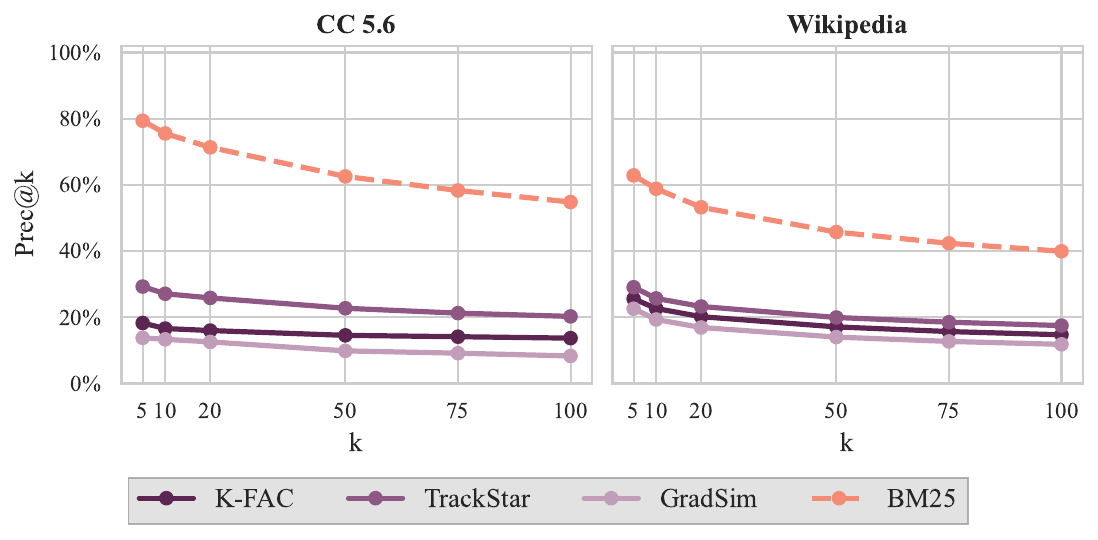}
\caption{\textbf{Precision@k of correct only ranking $\mathcal{I}(z, s^+_j)$ for facts}, across attribution methods and both training corpora (Common Corpus 5.6B and Wikipedia).}
\label{fig:prec_k_full_facts_good_only}
\end{figure*}

\begin{figure*}[!h]
    \centering
    \includegraphics[width=1\linewidth]{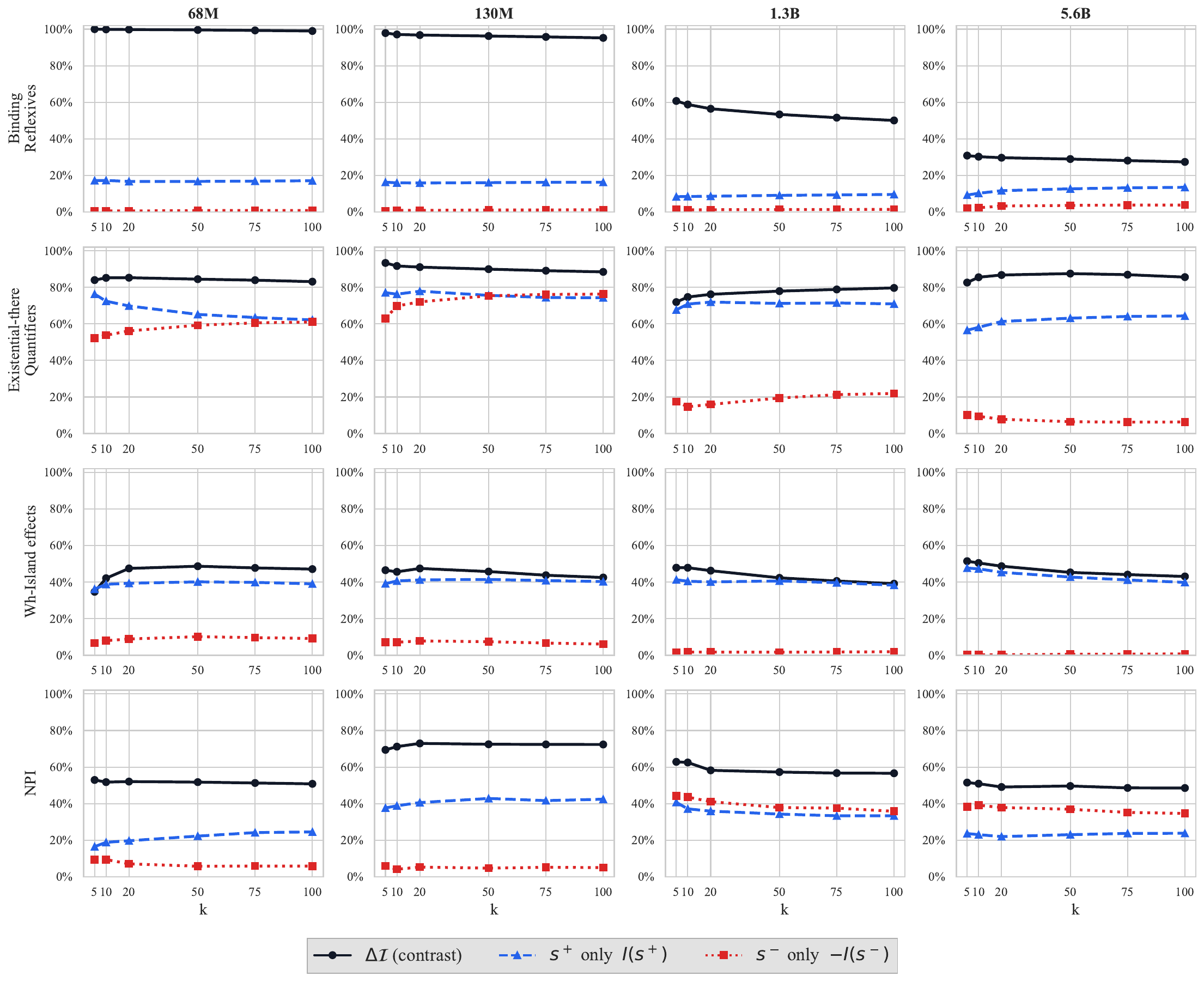}
    \caption{\textbf{Precision@k for the K-FAC approximation under three rankings: the influence contrast $\Delta \mathcal{I}$ (black, solid), ranking on the correct member alone, $\mathcal{I}(s^+)$ (blue, dashed), and ranking on the negated incorrect member alone, $-\mathcal{I}(s^-)$ (red, dotted), where $\Delta \mathcal{I} = \mathcal{I}(s^+) - \mathcal{I}(s^-)$.}
    For \textit{existential there quantifiers} and \textit{wh-island effects}, the contrast tracks the $\mathcal{I}(s^+)$ ranking fairly closely, indicating that the score is driven mostly by the correct member. For \textit{binding reflexives} and \textit{NPIs}, the contrast yields substantially higher Precision@k than either member alone, indicating that only the contrast carries enough signal to separate $\mathcal{D}_X$ from the base corpus: ranking on $\mathcal{I}(s^+)$ or $-\mathcal{I}(s^-)$ alone would in these cases be insufficient, so a non-contrastive view would lead to a biased interpretation of IFs. In both regimes, however, $-\mathcal{I}(s^-)$ contributes less than $\mathcal{I}(s^+)$: the correct member is the single well-defined target, whereas the incorrect member is only one representative of many possible wrong forms (e.g., \textit{himself} vs.\ \textit{herself}, \textit{themselves}), so on its own it is a weaker attribution signal.}
        \label{fig:contrast_good_bad_linguistics_kfac}
\end{figure*}

\begin{figure*}[!h]
    \centering
    \includegraphics[width=\linewidth]{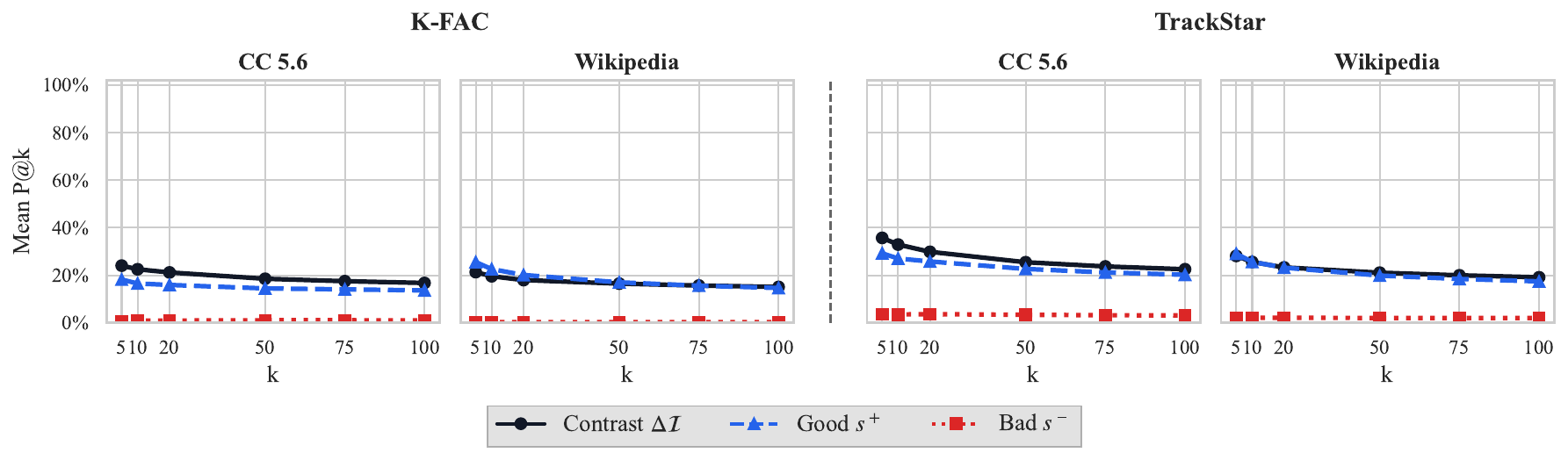}
\caption{\textbf{Precision@k for the K-FAC and TrackStar approximations on facts, under three rankings: the influence contrast $\Delta \mathcal{I}$ (black, solid), ranking on the correct member alone, $\mathcal{I}(s^+)$ (blue, dashed), and ranking on the negated incorrect member alone, $-\mathcal{I}(s^-)$ (red, dotted), where $\Delta \mathcal{I} = \mathcal{I}(s^+) - \mathcal{I}(s^-)$}, shown for the Common Corpus 5.6B and Wikipedia models. For both methods the contrast tracks the $\mathcal{I}(s^+)$ ranking closely, indicating that the score is driven mostly by the correct member. This is expected: the correct object is the single target, supported by the co-occurrence sentences in $\mathcal{D}_X$, whereas each incorrect object is only one of many candidate distractors with no dedicated supporting evidence, so on its own $-\mathcal{I}(s^-)$ is a weaker attribution signal.}
\label{fig:contrast_good_bad_facts_kfac}
\end{figure*}

\clearpage

\end{document}